\documentclass{article} 

\usepackage{natbib}
\usepackage[margin=0.9in]{geometry}

\usepackage{hyperref}
\usepackage{url}

\usepackage{microtype}
\usepackage{graphicx}
\usepackage{subfigure}
\usepackage{booktabs} 

\usepackage{amsmath}
\usepackage{amssymb}
\usepackage{mathtools}
\usepackage{amsthm}
\usepackage{bbm}

\usepackage{enumitem}
\setlist[itemize]{noitemsep, nolistsep, leftmargin=*}
\setlist[enumerate]{noitemsep, nolistsep, leftmargin=*}

\usepackage[capitalize,noabbrev]{cleveref}

\usepackage[toc,page,header]{appendix}
\usepackage{minitoc}

\usepackage{amsmath}
\usepackage{amssymb}
\usepackage{mathtools,comment}
\usepackage{amsthm}

\usepackage[capitalize,noabbrev]{cleveref}

\theoremstyle{plain}
\newtheorem{theorem}{Theorem}[section]

\newtheorem{lemma}[theorem]{Lemma}

\theoremstyle{definition}
\newtheorem{definition}[theorem]{Definition}

\theoremstyle{remark}

\newcommand{\generror}[2]{\mathrm{gen}(#1, #2)}

\newcommand{\R}{\mathbb{R}}

\newcommand{\ie}{\textit{i.e.}}

\usepackage[toc,page,header]{appendix}
\usepackage{minitoc}

\usepackage{hyperref}       
\usepackage{url}            
\usepackage{booktabs}       
\usepackage{amsfonts}       
\usepackage{nicefrac}       
\usepackage{microtype}      
\usepackage{xcolor}     
 
\usepackage{mathtools}
\usepackage{amssymb}
\usepackage{amsthm}
\usepackage{amsfonts}
\usepackage{bm}
\usepackage{bbm}

\usepackage{amsmath}

\usepackage{adjustbox}
\usepackage{multirow}
\usepackage{subfigure}
\usepackage{wrapfig}
\usepackage{placeins}
\usepackage{float}

\usepackage{algorithm}

\usepackage{algorithmic}

\usepackage{amsmath}
\usepackage{mathtools}
\usepackage{dsfont}
\usepackage{caption,subfigure,graphicx,epstopdf,multirow,multicol,booktabs,verbatim,wrapfig,bm}
\usepackage{makecell}
\usepackage{amsthm,amssymb,amsfonts,amsmath}
\usepackage{algorithmic}
\usepackage{subfiles}

\title{Asymmetry in Low-Rank Adapters of \\ Foundation Models}

\date{}

\def\month{MM}  
\def\year{YYYY} 
\def\openreview{\url{https://openreview.net/forum?id=XXXX}} 

\begin{document}

\maketitle

\vspace*{-2cm}

\begin{center}
{ 
\begin{tabular}{ccc}
Jiacheng Zhu$^{1}$ &  Kristjan Greenewald$^{3}$ & Kimia Nadjahi$^{1}$  \\
\end{tabular}

\begin{tabular}{ccc}
 Haitz S\'aez de Oc\'ariz Borde$^2$ & Rickard Br\"uel Gabrielsson$^1$ &  Leshem Choshen$^{1,3}$  \\
\end{tabular}

\begin{tabular}{ccc}
Marzyeh Ghassemi$^{1}$ & Mikhail Yurochkin$^3$ & Justin Solomon$^{1}$ \\
\end{tabular}
}


\vspace*{.1in}

{
\begin{tabular}{c}
MIT CSAIL$^1$, University of Oxford$^2$, MIT-IBM Watson AI Lab$^3$ \\
\end{tabular}
}
\vspace*{.2in}
\end{center}

\begin{abstract}

Parameter-efficient fine-tuning optimizes large, pre-trained foundation models by updating a subset of parameters; in this class, Low-Rank Adaptation (LoRA) is particularly effective. Inspired by an effort to investigate the different roles of LoRA matrices during fine-tuning, this paper characterizes and leverages unexpected asymmetry in the importance of low-rank adapter matrices. Specifically, when updating the parameter matrices of a neural network by adding a product $BA$, we observe that the $B$ and $A$ matrices have distinct functions: $A$ extracts features from the input, while $B$ uses these features to create the desired output. Based on this observation, we demonstrate that fine-tuning $B$ is inherently more effective than fine-tuning $A$, and that a random untrained $A$ should perform nearly as well as a fine-tuned one. Using an information-theoretic lens, we also bound the generalization of low-rank adapters, showing that the parameter savings of exclusively training $B$ improves the bound. We support our conclusions with experiments on RoBERTa, {BART-Large}, LLaMA-2, and ViTs. 
The code and data is available at \url{https://github.com/Jiacheng-Zhu-AIML/AsymmetryLoRA}
\end{abstract}

\doparttoc 
\faketableofcontents

\section{Introduction}

Foundation models for data-rich modalities such as text and imagery have achieved significant success by pre-training large models on vast amounts of data. While these models are designed to be general-purpose, it is often necessary to \emph{fine-tune} them for downstream tasks. However, the huge size of foundation models can make fine-tuning the entire model impossible, inspiring parameter-efficient fine-tuning (PEFT) methods that selectively update fewer parameters~\citep[c.f.][]{lialin2023scaling}. The effectiveness of PEFT demonstrates that updating even a tiny fraction of the parameters can retain and enrich the capabilities of pretrained models. Indeed, fine-tuning has become a necessary ingredient of modern ML; for example, the PEFT package~\citep{huggingface_peft} has supported more than 4.4k projects since its creation in November 2022.

Among PEFT methods, low-rank adaptation~(LoRA)~\citep{Hu2021LoRALA} has become increasingly popular, which leverages the assumption that over-parameterized models have a low intrinsic dimension~\citep{Aghajanyan_2021_intrinsic}. 
To update a neural network, LoRA trains a subset of the parameters (usually attention) by representing weight matrices as $W_0 + \Delta W$, where $W_0$ is the fixed weight matrix from the pre-trained model and $\Delta W$ is a low-rank update. Compared to full fine-tuning, LoRA considerably reduces the number of trainable parameters and memory requirements and often achieves similar or better performance. 


Most LoRA implementations factor $\Delta W = BA$ and optimize for $A$ and $B$, where $A$ and $B$ have fewer rows and columns (resp.) than $\Delta W$; this approach was proposed by \citet{Hu2021LoRALA}. With this set of variables, the standard LoRA training procedure—where $A$ is initialized to a random matrix and $B$ is initialized to zero—exhibits an interesting asymmetry, which is leveraged in some empirical follow-ups~\citep{zhang2023lorafa,kopiczko2024vera}. In particular, while training $B$ is critical for the performance of LoRA, even a \emph{randomly} initialized $A$ seems to suffice for strong performance. On the other hand, reversing the roles of $A$ and $B$ substantially decreases performance.

Delving into this empirical suggestion from prior work, this paper demonstrates that LoRA's components are inherently asymmetric. In fact, the asymmetry occurs even for linear models (\S\ref{sec:linear}). 
Indeed, our theoretical (\S\ref{sec:Analysis}) and empirical analysis (\S\ref{sec:exps}) suggests that fixing $A$ to a random orthogonal matrix can yield similar performance to full LoRA training, and that this adjustment may even promote generalization.  This observation is backed by a comprehensive empirical study, leading to practical suggestions for improving parameter efficiency and generalization of LoRA models. Our contributions are as follows:

\begin{figure*}
    \centering
    \subfigure[Random initialization, same task]{\includegraphics[width=0.31\textwidth]{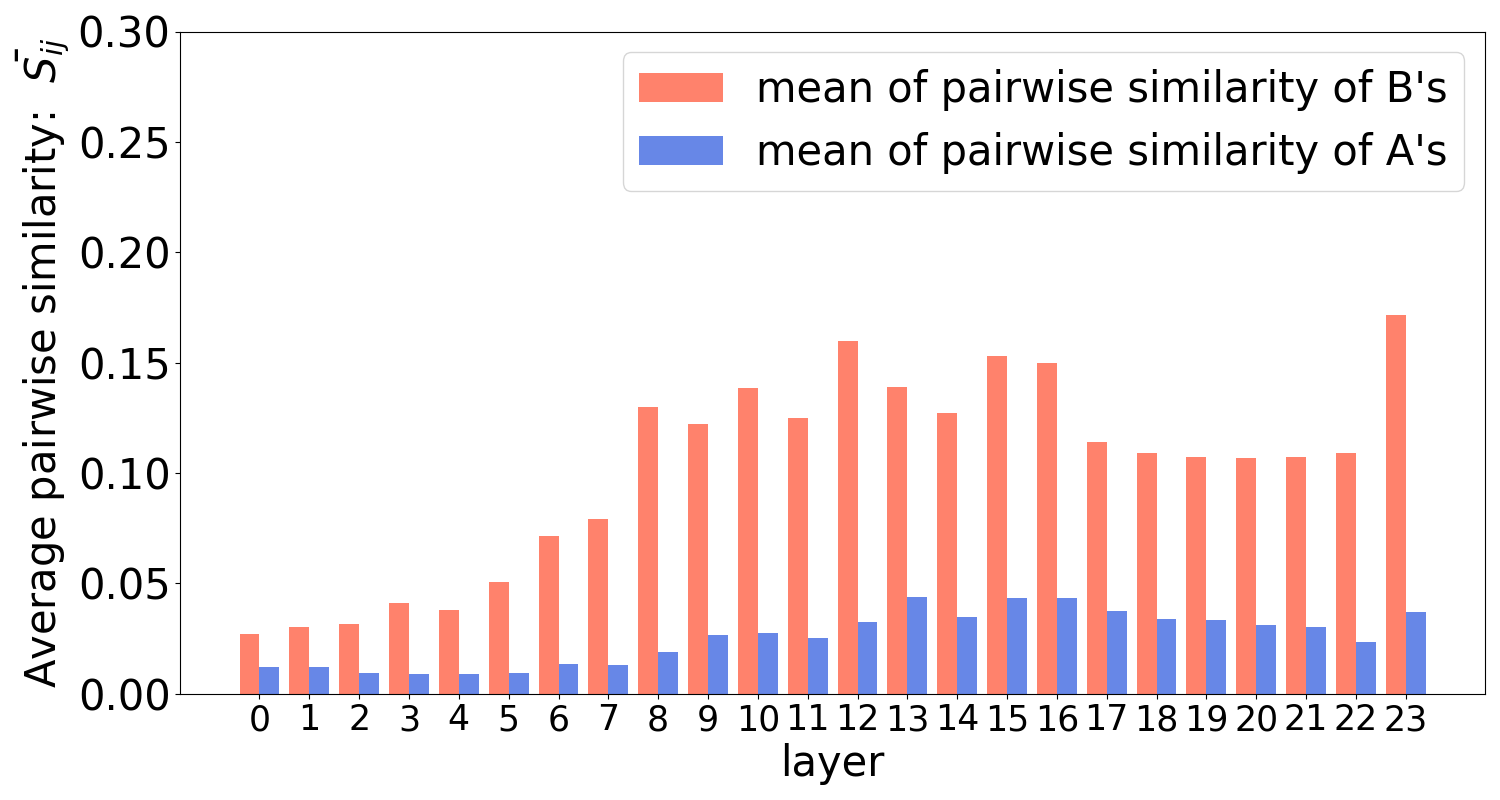} \label{fig1:same_task}
    } 
    \subfigure[Fixed initialization, different tasks
    ]{\includegraphics[width=0.31\textwidth]{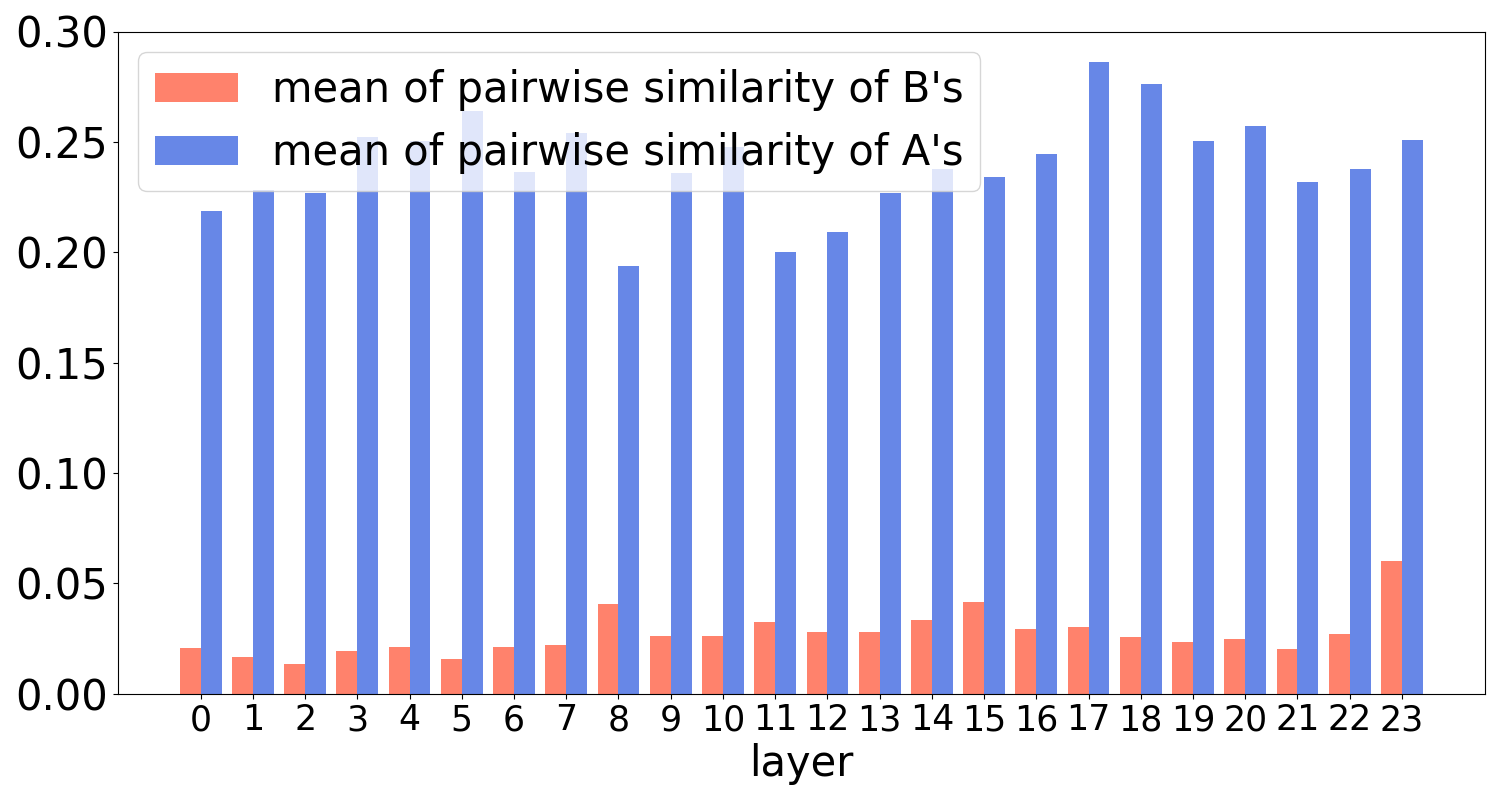}
    \label{fig1:diff_task}} 
    \subfigure[Random initialization, different tasks]{\includegraphics[width=0.31\textwidth]{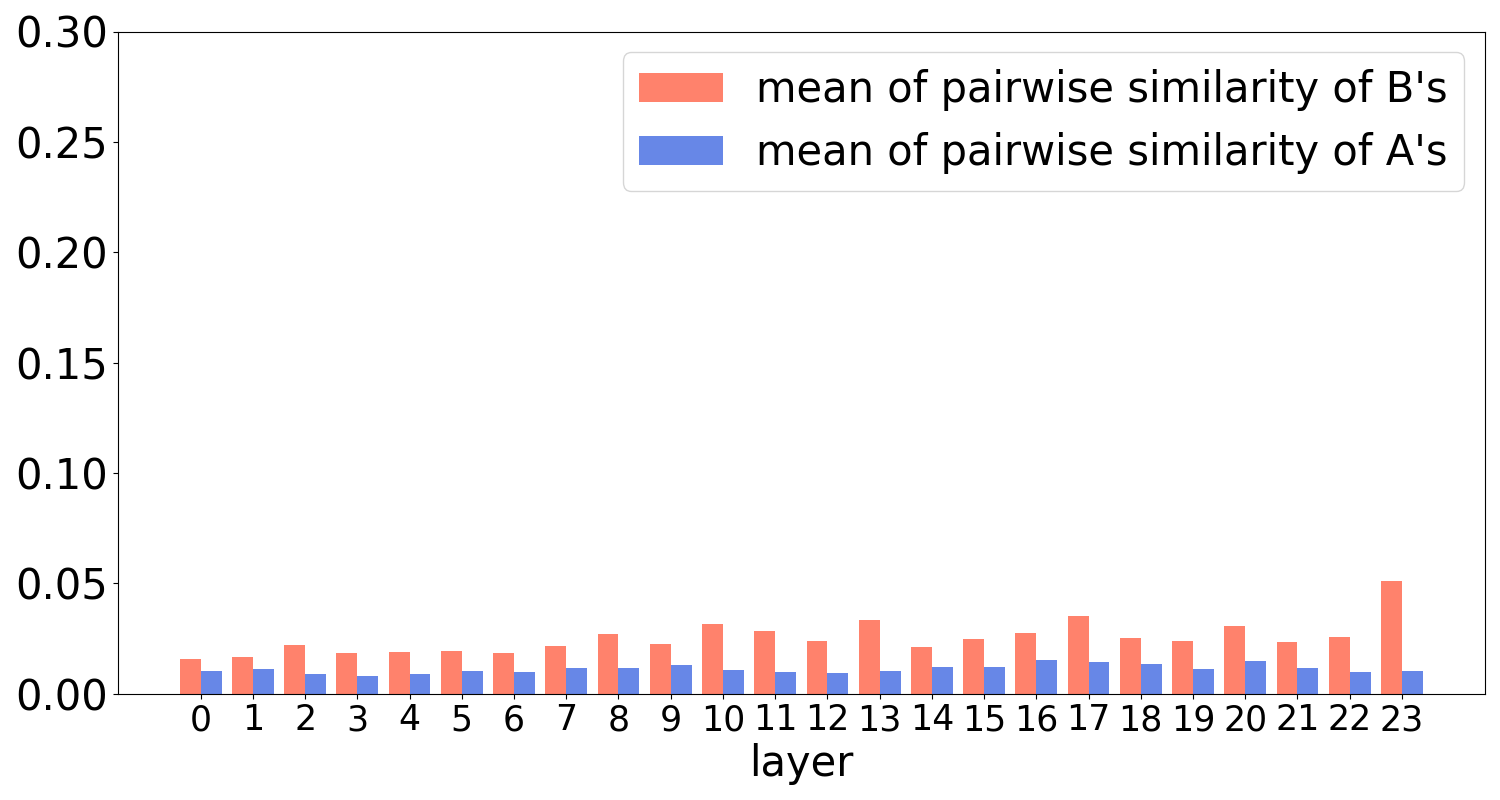} \label{fig1:sametask_diffseed}
    } 
    \vspace{-5pt}
    \caption{Similarity of learned LoRA matrices $A$ \& $B$ across layers of a RoBERTa model fine-tuned with different initialization and data settings. $B$s are similar when fine-tuning on the same task (a) and dissimilar when fine-tuning on different tasks (b and c). $A$s are similar when initialized identically (b), even though fine-tuning is done on different tasks, and dissimilar when initialized randomly regardless of the fine-tuning task (a and c). The experiment demonstrates the asymmetric roles of $A$ and $B$ in LoRA.
}
    \label{fig1}
\end{figure*}

\begin{itemize}
\item We provide simple theoretical and empirical analysis demonstrating \textit{asymmetry} of training the two adapter matrices, showing that tuning $B$ is more impactful than tuning $A$. This confirms and builds upon prior empirical observations~\citep{zhang2023lorafa,kopiczko2024vera}.
\item We show theoretically and empirically that randomly drawing and freezing $A$ while tuning only $B$ can improve generalization vs.\ tuning both $B$ and $A$, in addition to practical gains achieved by $2\times$ parameter reduction.
\item We validate our findings through experiments using models including RoBERTa, {BART-Large}, LLaMA-2, and the vision transformer (ViT), on both text and image datasets. 
\end{itemize}


\section{Related Work}


Since the introduction of the original LoRA technique~\citep{Hu2021LoRALA}, numerous enhancements have been proposed. For example, quantization can reduce memory usage during training~\citep{gholami2021survey_quantization,Dettmers2023QLoRAEF,guo2024lqlora}. 
Also, the number of trainable parameters can be further reduced by adaptively allocating the rank~\citep{zhang2023adalora}, pruning during training~\citep{Benedek2024PRILoRAPA}, or pruning and quantizing after training~\citep{yadav2023compeft}.

To further reduce the number of trainable LoRA parameters, the idea of reusing (randomly generated) weights or projections~\citep{frankle2018lottery,Ramanujan_2020_random_iccv} suggests strategies from learning diagonal matrices rescaling randomly-drawn and frozen $B,A$ matrices (VeRA)~\citep{kopiczko2024vera}, deriving $B$ and $A$ from the SVD decomposition of the pre-trained $W_0$ and optimizing for a smaller matrix in the resulting basis (SVDiff)~\citep{han2023svdiff}, learning a linear combination of fixed random matrices (NOLA)~\citep{koohpayegani2023nola}, or fine-tuning using orthogonal matrices (BOFT)~\citep{liu2024boft}. As echoed in our empirical results, previous methods observe that freezing $A$ in conventional LoRA preserves performance~\citep{zhang2023lorafa}. While nearly \emph{all} recent studies treat the two matrices asymmetrically in their initialization or freezing schemes, there is a lack of formal investigation into this asymmetry in low-rank adaptation.

\citet{zeng2023expressive} specifically investigate the expressive power of LoRA, but only focus on linearized networks and linear components. Their analysis does not consider aspects such as the particular distribution of the fine-tuning target data, generalization, or the differing roles of the different matrices. Lastly, we would like to highlight that even before LoRA, the effectiveness of fine-tuning was also explained by leveraging similar ideas related to the intrinsic low dimensionality of large models~\citep{Aghajanyan_2021_intrinsic}.

\section{Preliminaries \& Background}


\textbf{Notation.} Suppose we are given a pre-trained weight matrix $W_0 \in \mathbb{R}^{d_{\text{out}} \times d_{\text{in}}}$ representing a dense multiplication layer of a neural network foundation model. LoRA fine-tunes by updating the weights to $W_0 + \Delta W$, where $\text{rank}(\Delta W) = r \leq \min(d_{\text{out}}, d_{\text{in}})$. In particular, \citet{Hu2021LoRALA} factor $\Delta W = BA$, where $A \in \mathbb{R}^{r \times d_{\text{in}}}$ and $B \in \mathbb{R}^{d_{\text{out}} \times r}$ have restricted rank $\leq r$. During training, $W_0$ is fixed; LoRA updates $(A, B)$. This yields more efficient updates than full fine-tuning, provided that $r < \frac{d_{\text{in}} d_{\text{out}}}{d_\text{in} + d_{\text{out}}}$.

Now using $i$ to index layers of a network, a LoRA update is thus characterized by a set of pre-trained weight matrices $\mathbf{W} \triangleq \{W_i\}_{i=1}^L$, a set of pre-trained bias vectors $\mathbf{b} \triangleq \{b_i\}_{i=1}^L$, and a set of low-rank trainable weights $\Delta \mathbf{W} \triangleq \{\Delta W_i\}_{i=1}^{L'}$. LoRA may not update all $L$ weight matrices in $\mathbf{W}$, in which case $L' \leq L$.

\textbf{Motivating example.} In Figure \ref{fig1}, we investigate the similarity of learned matrices $A$ and $B$ under three scenarios:
\begin{enumerate}[label=(\alph*)]
\item random initialization, $A$ \& $B$ trained multiple times on the same task;\label{randominitsametask} 
\item fixed initialization, $A$ \& $B$ trained multiple times, each time on a different task; and \label{fixedinitdifferenttask}
\item random initialization, $A$ \& $B$ trained multiple times, each time on a different task. \label{randominitdifferenttask}
\end{enumerate}
Here, we fine-tune RoBERTa large \citep{liu2019roberta} with LoRA on the tasks from the GLUE benchmark~\citep{Wang_2018_glue}. Specifically, we fine-tuned \textit{mrpc} with 5 random seeds for (a) and on \textit{mrpc, rte, stsb, and cola} for (b) and (c). 

The figure plots similarity of learned $A$ and $B$ matrices across layers in Figure \ref{fig1}, measured by canonical correlation analysis goodness of fit \citep{ramsay1984matrix}; see Appendix \ref{app:fig1} for motivation.

These plots suggest that $B$ is predominantly responsible for learning, while $A$ is less important. Specifically, when training on the same task with different initializations (scenario \ref{randominitsametask}), the learned $B$ matrices are similar to each other, while when training on different tasks (scenarios \ref{fixedinitdifferenttask} and \ref{randominitdifferenttask}), they are different. On the contrary, the similarity of learned $A$ matrices is insensitive to training data and is determined by initialization; it is highest in scenario \ref{fixedinitdifferenttask} when the initialization is fixed even though training data differs. See Appendix \ref{app:fig1} for additional details of this experiment.


\section{Theoretical Analysis}
\label{sec:Analysis}

In this section, we analyze the asymmetry in prediction tasks and its effect on generalization. We discuss a general case rather than a specific neural network architecture, considering rank $r$ adaptation of any parameter matrix $W = W_0 + B A$ used multiplicatively on some input-dependent vector, i.e.,
\begin{equation}
\mathrm{layerOutput} = \psi((W_0 + BA) \cdot \phi(\mathrm{layerInput}), \dots)
\label{eq:Lora}
\end{equation}
for some differentiable functions $\psi, \phi$. Here, $\psi$ may take more arguments depending on $\mathrm{layerInput}$, which  may have their own low rank adapted parameter matrices.  This generic form encompasses both feedforward and attention layers. 

In this setting, $A$ serves to extract $r$ features from $\phi(\mathrm{layerInput})$, which are then used by $B$ to predict some desired output for future layers.  We will argue that training $B$ to predict the output is crucial for correct outputs, while using a random $A$ is often sufficient, as $B$ can be optimized to use whatever information is retained in the $r$-dimensional projection $A\cdot\phi(\mathrm{layerInput})$.

\subsection{$A$, $B$ asymmetry in prediction tasks}

If we wish to reduce the effort of training both $A$ and $B$ in \eqref{eq:Lora}, in principle either $A$ could be frozen and $B$ tuned or $B$ frozen and $A$ tuned. As shown in \S\ref{sec:exps} and elsewhere, these two options are not empirically equivalent: It is best to freeze $A$ and tune $B$. In this section, we seek to understand the principle behind this asymmetry by theoretically analyzing the fine-tuning of a class of prediction models. We first build intuition with least-squares linear regression.


\subsubsection{Multivariate linear least-squares}\label{sec:linear}

 As a simple example analogous to a single network layer, we study $d_{in}$-to-$d_{out}$ least-squares linear regression (in \eqref{eq:Lora}, set $\phi$, $\psi$ to be identity). Specifically, suppose there is an input $X \in \mathbb{R}^{d_{in}}$, an output $Y \in \mathbb{R}^{d_{out}}$, and a pre-trained linear model 
\[
y_{pre}(X) = W_0 X + b_0,
\]
where $W_0 \in \mathbb{R}^{d_{out} \times d_{in}}$ and $b_0 \in \mathbb{R}^{d_{out}}$. 
With this model held constant, our goal is regressing $(Y_{targ},X_{targ})$ pairs where $Y_{targ}$ is given by:
\begin{align*}
&Y_{targ} = W_{targ} X_{targ} + b_{targ} 
\end{align*}
with $W_{targ} = W_0 + \Delta$. Following LoRA, we model the target $\Delta$ using a low rank update to the pre-trained $W_0$, i.e. $W = W_0 + BA$: 
\[
\hat{y}(x) = (W_0 + BA) x + b,
\]
where $B \in \mathbb{R}^{d_{out} \times r}$ and $A \in \mathbb{R}^{r \times d_{in} }$ for some $r$. 

To find an $A$ and $B$ that best matches the output, we optimize the least squares loss on the difference between $\hat{y}$ and $Y_{targ}$:
\begin{equation}
\mathcal{L}(A,B) \!=\! \mathbb{E}_{(Y_{targ}, X_{targ})} [\left\| Y_{targ} \!-\! (W_0 \!+\! BA) X_{targ} \!-\! b\right\|_2^2].
\label{eq:L2loss}
\end{equation}

Below, we present lemmas on minimizing this loss while freezing either $A$ or $B$. In both, for simplicity, we set $b = b_{targ}$ and $\mathbb{E}[X_{targ}] = 0$ and defer proofs to Appendix \ref{app:LS}.
\begin{lemma}[Freezing $A$ yields regression on projected \textbf{features}]
    \label{thm:A_L2}
    Optimizing $\mathcal{L}(A,B)$ while fixing $A = Q$ with $Q Q^\top = I_r$ yields 
    \[
        B^\ast = \Delta \Sigma Q^\top  (Q \Sigma Q^\top )^{-1},
    \]
    where $\Sigma = \mathrm{Cov}[ X_{targ} ]$, with expected loss 
    \begin{align*}
 \mathcal{L}(Q,B^{\ast})  =\ d_{out} \sigma^2 + \mathrm{Tr}[\Delta \Sigma \Delta^\top ] - 
 \mathrm{Tr} [Q \Sigma \Delta^\top  \Delta \Sigma Q^\top  (Q \Sigma Q^\top )^{-1}].
\end{align*}
\end{lemma}
\begin{lemma}[Freezing $B$ yields regression on projected \textbf{outputs}]
\label{thm:B_L2}
    Optimizing $\mathcal{L}(A,B)$ while fixing $B = U$ with $U^\top U = I_r$ 
    yields
    \[
    A^\ast = U^\top  (W_{targ} - W_0),
    \]
    with expected loss
    \begin{align*}
    \mathcal{L}(A^\ast,U) = d_{out} \sigma^2 + \mathrm{Tr}[\Delta \Sigma \Delta^\top ] - \mathrm{Tr}[U^\top  \Delta \Sigma \Delta^\top  U],
\end{align*}
where $\Sigma = \mathrm{Cov}[ X_{targ} ]$.
\end{lemma}
Comparing the lemmas above, $A^\ast$ is simply the $U$ projection of the targeted change in weight matrix $\Delta = W_{targ} - W_0$. Unlike $B^\ast$, the optimal choice of $A^\ast$ does not consider the input data distribution captured by $\Sigma$. 

Intuitively, if the goal of adaptation is to approximate some desired output, then projecting away the majority (since $r \ll d_{out}$) of the output is undesirable.  
In contrast, projecting away a portion of the input feature space will be less damaging, if the information $X_{targ}$ contains about $Y_{targ}$ is redundant (c.f., neuron dropout \citep{srivastava2014dropout} in neural network training) or if the distribution of $X_{targ}$ tends to be low-rank. 

Consider the following extreme example. If $\Sigma = F F^\top $ is at most rank $r$, e.g.\ if $F\in {d_{in} \times r}$, then for each $X$ there exists\footnote{Here $F^{\dag}$ denotes pseudoinverse.} an $N = F^{\dag} X\in \mathbb{R}^r$ such that $X = F N$. Suppose you have tuned a pair $A_\ast$, $B_\ast$. For any orthonormal $Q\in\mathbb{R}^{r \times d_{in}}$ (e.g. one drawn at random), we can write
\begin{align*}
    B_\ast A_\ast X = B_\ast A_\ast F N 
    = (B_\ast A_\ast F (QF)^{-1}) Q X,
\end{align*}
i.e. regardless of $A_\ast$, $B_\ast$, for any (random) $Q$, there is an exactly equivalent LoRA adaptation with $A = Q$ and $B = (B_\ast A_\ast F (QF)^{-1})$.
 In this setting, therefore, randomizing $A$ (to $Q$) is equally expressive to tuning it (using $A_\ast$). 


This intuition is also reflected in the typical LoRA initialization. When doing full LoRA (tuning both $A,B$), $A$ usually is initialized to a random Gaussian matrix, and $B$ is initialized to zero. This procedure---presumably empirically derived by \citet{Hu2021LoRALA}---intuitively fits our analysis above, since random $A$ yields good random predictive features, in contrast to using a random output prediction basis. Initializing $B$ to zero then starts the optimization at a zero perturbation of the pretrained model. 

We validate the above intuition with the following theorem:
\begin{theorem}[$A$, $B$ output fit asymmetry]
\label{thm:ABcomp}
Consider the settings of Lemmas \ref{thm:A_L2} and \ref{thm:B_L2}, and suppose $U, Q$ are sampled uniformly from their respective Stiefel manifolds. Then, 
$
\mathcal{L}(A^\ast, U) \geq \mathcal{L}(Q, B^\ast)
$
with high probability as $\nicefrac dr \rightarrow \infty$.
\end{theorem}
In other words, the least-squares prediction loss of only fine-tuning $B$ is at least as good as only fine-tuning $A$. 

\textbf{Intuition on asymmetry gap.} Theorem \ref{thm:ABcomp} is built on the following inequality:
\begin{align}
\mathrm{Tr}[\Sigma Q^\top  (Q \Sigma Q^\top )^{-1} Q \Sigma \Delta^\top  \Delta] \nonumber \geq \mathrm{Tr}[(Q^\top  Q) \Sigma Q^\top  (Q \Sigma Q^\top )^{-1} Q \Sigma \Delta^\top  \Delta].
\label{eq:ineq}
\end{align}
Let us consider an example regime to build intuition on the size of this gap. Following intuition that freezing $A$ is most successful when the information content of the input is redundant (c.f., \citet{Aghajanyan_2021_intrinsic}), suppose the distribution of $X$ is low rank, i.e., $\Sigma$ is of rank $r_X$. We can then write $\Sigma = U_X S_X U_X^\top $, where $U_X\in \mathbb{R}^{d_{in} \times r_X}$ is orthogonal and $S_X \in \mathbb{R}^{r_X \times r_X}$ is diagonal with nonnegative real entries.

For intuition, set $r_X = r$ and $S_X = \sigma^2 I_{r}$. We then have 
\begin{align*}
\Sigma Q^\top  (Q \Sigma Q^\top )^{-1} Q \Sigma \Delta^\top  \Delta 
= \sigma^2 U_X U_X^\top  \Delta^\top  \Delta,
\end{align*}
which no longer depends on $Q$. The expectation of the key inequality gap in \eqref{eq:ineq} then becomes 
\begin{align*}
    &\mathbb{E}_Q\mathrm{Tr}[\Sigma Q^\top  (Q \Sigma Q^\top )^{-1} Q \Sigma \Delta^\top  \Delta]
    - \mathbb{E}_Q\mathrm{Tr}[(Q^\top  Q) \Sigma Q^\top  (Q \Sigma Q^\top )^{-1} Q \Sigma \Delta^\top  \Delta]\\
    &= \mathbb{E}_Q\mathrm{Tr}[(I - Q^\top  Q)\sigma^2 U_X U_X^\top  \Delta^\top  \Delta ]
    {} \rightarrow \left(1 - \frac{r}{d}\right) \mathrm{Tr}[U_X U_X^\top  \Delta^\top  \Delta]
\end{align*}
as $d$ becomes large. In other words, the performance advantage of tuning $B$ over $A$ is large when $d \gg r$, which is the typical regime in practice.

\subsubsection{Nonlinear losses and multilayer models}
Recalling \eqref{eq:Lora} with an input transformation $\phi$ and output transformation $\psi$, consider losses on the output of the form
\begin{equation} \label{eq:genreg_lkl}
    \mathcal{L}(W) = \sum_{i=1}^n h(f(\psi(W \phi(x_i)))) - y_i^\top  f(\psi(W \phi(x_i))),
\end{equation}
where $f, h$ are differentiable functions specified by the desired loss, $y_i \in \mathbb{R}^K$, $x_i \in \mathbb{R}^{d_{in}}$, and $W \in \mathbb{R}^{d_{out} \times d_{in}}$. This class contains logistic regression (with $y$ being a one-hot encoded class vector), least-squares regression, and generalized linear regression---including a neural network with cross entropy loss with one layer being tuned.

We next analyze the gradient of this loss. Our argument is stated with one adapted parameter matrix, but it directly applicable to multilayer and transformer networks with multiple matrices being adapted, where $\phi$, $\psi$, and $f$ will in that scenario vary depending on each parameter matrix's position in the network; $\phi$, $\psi$, and $f$ will depend on other parameter matrices and the current value of their adaptations (by definition of gradients). The interpretation will now be that fixing $A$ when adapting a parameter matrix $W^{(\ell)}$ projects the inputs of the corresponding parameter matrix to a lower-dimensional subspace while retaining the ability to fully match the outputs, and fixing $B$ correspondingly projects the parameter matrix's outputs.

For simplicity of notation, the remaining derivation in this section takes $\phi,\psi$ to be the identity; the extension to general $\phi,\psi$ is clear. 
%
%
Then, the gradient of \eqref{eq:genreg_lkl} is
\begin{equation}
\label{eq:Wgen}
    \nabla_W  \mathcal{L}(W) = \sum_{i=1}^n J^\top _f(W x_i)\left[\nabla h(f(W x_i)) - y_i\right] x_i^\top  ,
\end{equation}
where $J_f$ is the Jacobian of $f$. 
Starting from this formula, below we incorporate \eqref{eq:Lora} by taking $W = W_0 + B A.$

\textbf{Freezing $A$.} Freezing $A = Q$ yields 
\begin{equation}
\label{eq:Agen}
\nonumber \nabla_B  \mathcal{L}(BQ + W_0)= \sum_{i=1}^n\! J^\top _f\!((BQ + W_0) x_i)\left[\nabla h(f((W_0 + BQ) x_i)) \!-\! y_i\right] (Q x_i)^\top. 
\end{equation}
Like the least-squares case, the input data is projected by $Q$ but the output $y_i$ is unaffected. 

\textbf{Freezing $B$.} Freezing $B = U$ yields 
\begin{equation}
\label{eq:Bgen}
\nonumber \nabla_A  \mathcal{L}(UA + W_0) = U^\top \! \sum_{i=1}^n\! J^\top _f((UA + W_0) x_i)\left[\nabla h(f((W_0 + UA) x_i))\! - \!y_i\right] x_i^\top .
\end{equation}
Here, the coefficient of $x_i^\top$
can be thought of as the output fit term. It includes the Jacobian of $f$ since $f$ is applied between the weights and the output. Compared to \eqref{eq:Wgen} and \eqref{eq:Agen}, in \eqref{eq:Bgen} this output fit term is projected  by $U$. If $f$ is (near) linear, 
then this projection will be (approximately) data-independent, highlighting the loss of output information when freezing $B$.

Hence, in this more general setting, the different roles of $A$ and $B$ are still apparent, and we expect an asymmetry in being able to fit the output. 

\textbf{Example: Logistic regression.} For multiclass logistic regression, we have a training dataset $\{(x_i, c_i)\}_{i=1}^n$ where $x_i \in \R^{d}$ (features) and $c_i \in \{1, \dots, K\}$ (label). Denote by $y_i \in \R^{K}$ the vector with $y_{c_i} = 1$ and $y_{k} = 0$ for $k \neq c_i$. The log likelihood is the cross-entropy error
\begin{equation} \label{eq:logreg_lkl}
    \mathcal{L}(w_1, \dots, w_K) = \sum_{i=1}^n \sum_{k=1}^K y_i \ln(p_{i,k}),
\end{equation}
where $p_{i,k} = \frac{\exp(w_k^\top  x_i)}{\sum_{l=1}^K \exp(w_l^\top  x_i)}$ and $w_k \in \R^d$. 
Let $W \in \R^{K \times d}$ whose $k$-th row is $w_k$. Then, \eqref{eq:logreg_lkl} becomes
\begin{align} \label{eq:logreg_lkl_2}
    \nonumber\mathcal{L}(W) 
    = \sum_{i=1}^n  \ln(\mathbf{1}^\top e^{W x_i})  - y_i^\top  W x_i,
\end{align}
where $\mathbf{1}$ is the column vector of size $K$ with all elements equal to 1; note $y_i^\top  \mathbf{1} = 1$ due to the one-hot structure. This loss can be put in the form \eqref{eq:genreg_lkl} by setting 
$f(z) = z$ and $h(z) = \ln(\mathbf{1}^\top e^{z})$. For freezing, we then have 
\begin{align*}
    \nabla_A  \mathcal{L}(UA) = U^\top  \sum_{i=1}^n (y_i - p_i(UA)) x_i^\top  \qquad \textrm{ and }  \qquad
{} \vspace{1cm} \text{ {}} 
    \nabla_B  \mathcal{L}(BQ) = \sum_{i=1}^n (y_i - p_i(BQ)) (Q x_i)^\top,
\end{align*}
where $p_i(W) =  \frac{e^{W x_i}}{\mathbf{1}^\top  e^{W x_i}} \in \R^K$.
Freezing $B = U$,  as in least-squares, implies that each output $y_i$ is projected as $U^\top  y_i$, implying that, at best, the model can hope to only learn outputs in the small random subspace $U$. In contrast, freezing $A=Q$ is  equivalent to logistic regression on the full output with features projected by $Q$: $\{(Qx_i, y_i)\}_{i=1}^n$. 










\subsection{Advantages of tuning only $B$ over $BA$ together}
\label{Advantages of tuning $B$ only over $BA$ together}

In the previous section, we established that fine-tuning $B$ alone is typically superior to fine-tuning $A$ alone. It remains, however, to motivate fine-tuning $B$ alone over fine-tuning both $A$ and $B$ together. In this section, we show that the reduced amount of adapted parameters by (roughly) half provides computational gains and improvements in information-theoretic generalization bounds. 

\subsubsection{Number of parameters} The key benefit of LoRA is parameter efficiency, which saves memory during training, storage and communication \cite{lialin2023scaling}. Fine-tuning $B$ alone as opposed to both $A$ and $B$ reduces the number of parameters by a factor of $\frac{d_{out} }{d_{out} + d_{in}}$, which equals $0.5$ when $d_{in} =d_{out}$. 

\subsubsection{Generalization bounds}

Consider a learning task, where the training examples lie in $\mathcal{Z} = \mathcal{X} \times \mathcal{Y}$; here, $\mathcal{X}$ denotes the feature space and $\mathcal{Y}$ is the label space. Suppose one observes a training set $S_n \triangleq (Z_1, \dots, Z_n) \in \mathcal{Z}^n$, with $n$ i.i.d.\ training examples from unknown distribution $\mu$. Denote by $\mu^{\otimes n} = \mu \times \dots \times \mu$ the distribution of $S_n$. The objective of the learner is to find a predictor $f : \mathcal{X} \to \mathcal{Y}$ that maps features to their labels. We assume each predictor is parameterized by $w \in \mathcal{W}$ (e.g., if $f$ is a neural network, $w$ denotes its weights). Denote by $\mathcal{A} : \mathcal{Z}^n \to \mathcal{W}$ the learning algorithm which selects a predictor given $S_n$. $\mathcal{A}$ is, in general, a probabilistic mapping, and we denote by $P_{W|S_n}$ the distribution of its output $W$ given input $S_n$. If $\ell : \mathcal{W} \times \mathcal{Z} \to \mathbb{R}_+$ is a loss, we define: 
\begin{align*}
\textrm{Population risk:}\quad &
\mathcal{R}_\mu(w) \triangleq \mathbb{E}_{Z \sim \mu}[\ell(w, Z)]
\\
\textrm{Empirical risk:}\quad &
\widehat{\mathcal{R}}_n(w) \triangleq \frac1n \sum_{i=1}^n \ell(w, Z_i).
\end{align*}
The generalization error of $\mathcal{A}$ is 
\begin{equation}
    \generror{\mu}{\mathcal{A}} \triangleq \mathbb{E}_{(W, S_n) \sim P_{W | S_n} \times \mu^{\otimes n}} [\mathcal{R}_\mu(W) - \widehat{\mathcal{R}}_n(W)] \,.
    \nonumber
\end{equation}
We bound this generalization error using the information-theoretic generalization framework of \citet{xu2017}. 
Consider the following incarnations of fine-tuning algorithms, corresponding to classic LoRA (tuning both $A, B$ matrices), tuning only $B$, and tuning only $A$:
\begin{definition}[Fine-tuning algorithms]
\label{def:alg}
Let $\mathbf{W} = \{W_i\}_{i=1}^L$ be the $L$ parameter matrices of a pretrained model. Let $\mathcal{I}\subseteq \{1,\dots, L\}$ be a specified subset of the parameter matrices to be fine-tuned. Given a fine-tuning training set $S_n$, let $r$ be a chosen rank and suppose each tuned parameter is quantized to $q$ bits. We define the following algorithmic frameworks (other details can be arbitrary) for choosing an adaptation $\mathbf{\Delta W} = \{\Delta_i\}_{i \in \mathcal{I}}$, yielding a fine-tuned $W_{tuned} = \{W_{tuned,i}\}_{i=1}^L$ with $W_{tuned,i} = W_i + \Delta_i$ for $i \in \mathcal{I}$ and $W_{tuned,i} = W_i$ otherwise:
\begin{itemize}
    \item $\mathcal{A}_{BA}$: For each $i \in \mathcal{I}$, constrain $\Delta_i = B_i A_i$ and optimize $\{B_i, A_i\}_{i \in \mathcal{I}}$ to fit the data $S_n$.
    \item $\mathcal{A}_{B}$: For each $i \in \mathcal{I}$, sample $Q_i\in \mathbb{R}^{r \times d^{(i)}_{in}}$ at random, constrain $\Delta_i = B_i Q_i$, and optimize $\{B_i\}_{i \in \mathcal{I}}$ to fit the data $S_n$.
    \item $\mathcal{A}_{A}$: For each $i \in \mathcal{I}$, sample $U_i\in \mathbb{R}^{d_{out}^{(i)} \times r}$ at random, constrain $\Delta_i = U_i A_i$, and optimize $\{A_i\}_{i \in \mathcal{I}}$ to fit the data $S_n$.
\end{itemize}
\end{definition}
We have the following lemma, proved in Appendix \ref{app:IT}:
\begin{lemma}[Generalization bounds on adapting $A$ and/or $B$]
\label{lemm:genbd}
    Consider the algorithms of Definition \ref{def:alg}. Assume that 
$\ell^{\mathbf{W}, \mathbf{b}}(\Delta \mathbf{W}, \widetilde{Z})$ is $\sigma$-sub-Gaussian\footnote{Bounded losses are sub-Gaussian.} under $(\Delta \mathbf{W}, \widetilde{Z}) \sim P_{\Delta \mathbf{W} | \mathbf{W}, \mathbf{b}} \times \mu$.
Then,
\begin{align*}
    | \generror{\mu}{\mathcal{A}_{BA}} | &\leq \sqrt{\frac{{2rq\sigma^2\ln 2}}{n}  \sum_{i\in \mathcal{I}} (d_{in}^{(i)} + d_{out}^{(i)} )},\\
    | \generror{\mu}{\mathcal{A}_{B}} | &\leq \sqrt{\frac{{2rq\sigma^2\ln 2}}{n}  \sum_{i\in \mathcal{I}} d_{out}^{(i)} },\\
    | \generror{\mu}{\mathcal{A}_{A}} | &\leq \sqrt{\frac{{2rq\sigma^2\ln 2}}{n}  \sum_{i\in \mathcal{I}} d_{in}^{(i)}}.
\end{align*}
\end{lemma}
This generalization bound increases with the number of parameters being tuned, which is an increasing function of $r$ and the dimensions of the parameter matrices. Importantly, since tuning just one factor ($A$ or $B$) involves tuning fewer parameters than $A$ and $B$ together, the generalization bound is correspondingly smaller. In the case where the $d_{in}^{(i)} = d_{out}^{(i)}$, the bound for tuning one factor only is a factor of $\sqrt{2}$ smaller than the bound for tuning both factors, implying that the rank $r$ for $\mathcal{A}_B$ could be doubled and have a generalization bound matching that of $\mathcal{A}_{BA}$.









\subsection{Discussion of theoretical analysis}
\label{theoretical_discussion}
The previous two sections establish two conclusions: (1) Tuning $A$ has limited importance when trying to match a desired output; and (2) Tuning one factor instead of two reduces the number of parameters for the same $r$, while improving generalization bounds and potentially providing memory benefits.

Given a fixed parameter count and generalization budget, therefore, we can use a larger $r = r_{B}$ when fine-tuning $B$ alone than the $r_{BA}$ that would be used on standard LoRA fine-tuning both $A$ and $B$. This addition provides more expressive power for the same number of parameters without loss of generalization bounds. Hence, when matching parameter or generalization budget, we expect that fine-tuning a rank-$r_B$ $B$ typically improves performance over fine-tuning a rank-$r_{BA}$ $BA$ LoRA adaptation. 

\begin{table*}[t!]
  \centering
  \addtolength{\tabcolsep}{-4pt}

  \caption{Different adaptation methods on the GLUE benchmark. We report the overall (matched and mismatched) accuracy for MNLI, Matthew's correlation coefficient for CoLA, Pearson correlation for STS-B, and accuracy for other tasks. Higher is better for all metrics. 
  }
  \label{tab:NLU_all_results}
    \resizebox{1.0\columnwidth}{!}{%
  \begin{tabular}{l|c|cccccccc}
  \hline
  \toprule
  Model \& Method & \# Trainable & \multicolumn{8}{c}{} \\
         & Parameters & MNLI & SST-2 & MRPC & CoLA & QNLI  & RTE & STS-B & Avg. \\
  \midrule
  \midrule
  LoRA $(r=8)$ & 0.8\% & 90.3\textsubscript{$\pm$.07}  & 95.6\textsubscript{$\pm$0.36} & {90.3}\textsubscript{$\pm$0.85} & 64.4\textsubscript{$\pm$1.8} & {94.0}\textsubscript{$\pm$0.29} &  {84.1}\textsubscript{$\pm$0.96}  & {91.5}\textsubscript{$\pm$0.16} & 87.2 \\
  
  AdaLoRA & 2.5\% & 90.4\textsubscript{$\pm$.37}  & 95.9\textsubscript{$\pm$.13} & 90.1\textsubscript{$\pm$.54} & {67.5}\textsubscript{$\pm$1.3} & {94.7}\textsubscript{$\pm$.22} & {85.4}\textsubscript{$\pm$.20} &  {91.3}\textsubscript{$\pm$1.0}  & 87.9 \\
  
  $(\text{IA})^3$ & 0.7\% & 90.0\textsubscript{$\pm$.21} & 95.4\textsubscript{$\pm$.17} & 83.7\textsubscript{$\pm$.13} & {57.6}\textsubscript{$\pm$.67} & {93.7}\textsubscript{$\pm$.07} & {70.3}\textsubscript{$\pm$1.5} & {87.0}\textsubscript{$\pm$0.4} & 82.5\\
  
  LoRA-FA & 0.3\% & 90.3\textsubscript{$\pm$.06} & 95.6\textsubscript{$\pm$.17} & 90.6\textsubscript{$\pm$.32} & {67.3}\textsubscript{$\pm$2.3}  & 93.4\textsubscript{$\pm$.61}  &   82.4\textsubscript{$\pm$1.4}  &  91.2\textsubscript{$\pm$.29} &  87.3 \\
  
  \midrule
  \midrule
  

  $\hat{\textbf{B}}_0 A_{rand}$ $(r=8)$ & 0.3\% & 90.1\textsubscript{$\pm$.19} &  {\underline{95.8}\textsubscript{$\pm$.29}} & 89.7\textsubscript{$\pm$.13} & \textbf{\underline{67.5}\textsubscript{$\pm$1.2}} & {\underline{94.0}\textsubscript{$\pm$.27}} &  82.8\textsubscript{$\pm$1.5}   & \textbf{\underline{91.9}\textsubscript{$\pm$.26}} & 87.4  \\

  $\hat{\textbf{B}}_0 A_{rand}$ $(r=16)$ & 0.8\% & 90.1\textsubscript{$\pm$.20} & \textbf{\underline{96.1}\textsubscript{$\pm$.18}} & \textbf{\underline{90.7}\textsubscript{$\pm$.90}} & \underline{66.1}\textsubscript{$\pm$2.6} & \textbf{\underline{94.4}\textsubscript{$\pm$.10}}  &   \textbf{\underline{84.1}\textsubscript{$\pm$.96}}  & 91.2\textsubscript{$\pm$.42}  & 87.5 \\
  

\cmidrule{3-10}

${B}_{rand} \hat{\textbf{A}}_0$ $(r=8)$ & 0.3\% & \textbf{\underline{90.3}\textsubscript{$\pm$.18}} & 95.5\textsubscript{$\pm$.66} & 89.3\textsubscript{$\pm$.09} &   {58.7}\textsubscript{$\pm$2.5} & {93.8}\textsubscript{$\pm$.21} &   {77.1}\textsubscript{$\pm$1.3} & 90.7\textsubscript{$\pm$.31} &  84.2 \\

 ${B}_{rand} \hat{\textbf{A}}_0$ $(r=16)$ & 0.8\% & 
  89.9\textsubscript{$\pm$.19}& \underline{95.6}\textsubscript{$\pm$.64} & 90.2\textsubscript{$\pm$0.23} &  60.3\textsubscript{$\pm$3.3} & 93.9\textsubscript{$\pm$0.25} &   {80.4}\textsubscript{$\pm$0.21} & 90.9\textsubscript{$\pm$0.13} & 85.9 \\
  

  
  \bottomrule
  \end{tabular}
  }
\end{table*}

\begin{table*}[t!]
  \centering
  \addtolength{\tabcolsep}{-4pt}
  
  \caption{ Different initialization of classic LoRA, setting either $A$ or $B$ to be zeros. Note that the trained result is not sensitive to different initializations, with performance differences tending to be smaller than the standard error.
  }
  \label{tab:NLU_initialization}
  \resizebox{1.0\columnwidth}{!}{%
  \begin{tabular}{l|c|ccccccccc}
  \hline
  \toprule
  Model \& Method & \# Trainable & \multicolumn{8}{c}{} \\
         & Parameters & MNLI & SST-2 & MRPC & CoLA & QNLI & RTE & STS-B & Avg. \\
  \midrule
  
  \midrule 

  $\hat{\mathbf B}_0 \hat{\mathbf A}_{V}$  & 0.8\% & \textbf{90.4\textsubscript{$\pm$0.11}} & 95.9\textsubscript{$\pm$0.16}  & 90.7\textsubscript{$\pm$0.84} & 64.0\textsubscript{$\pm$0.50} & 94.4\textsubscript{$\pm$0.16} &  {84.1}\textsubscript{$\pm$0.15} & 91.8\textsubscript{$\pm$00.15} & 87.3 \\
  
  $\hat{\mathbf B}_0 \hat{\mathbf A}_{rand}$  & 0.8\% & \textbf{90.4\textsubscript{$\pm$0.15}} & 96.0\textsubscript{$\pm$0.11} & 91.5\textsubscript{$\pm$1.1} & 64.1\textsubscript{$\pm$0.67} & 94.5\textsubscript{$\pm$0.11} &  \textbf{85.6\textsubscript{$\pm$0.96}} & \textbf{92.0\textsubscript{$\pm$0.31}} & 87.7 \\
  
  
  $\hat{\mathbf B}_U \hat{\mathbf A}_0$ & 0.8\% & 90.3\textsubscript{$\pm$0.07} & \textbf{96.1\textsubscript{$\pm$.18}} & \textbf{91.7\textsubscript{$\pm$0.33}} & 64.9\textsubscript{$\pm$1.5}  & \textbf{94.7\textsubscript{$\pm$0.33}}  &  {84.8}\textsubscript{$\pm$0.96} & 91.9\textsubscript{$\pm$0.19} & 87.8 \\
  
  $\hat{\mathbf B}_{rand} \hat{\mathbf A}_0$  & 0.8\% & 90.3\textsubscript{$\pm$0.27}  & 96.0\textsubscript{$\pm$.26} & 90.8\textsubscript{$\pm$0.51} & \textbf{66.0\textsubscript{$\pm$1.01}} & 94.5\textsubscript{$\pm$0.38} &  {83.6}\textsubscript{$\pm$1.5} & \textbf{92.0\textsubscript{$\pm$0.18}}  &  87.8 \\
  
  
  \bottomrule
  \end{tabular}}
\end{table*}

\section{Experiments}\label{sec:exps}

We investigate the asymmetry of low-rank adaptation methods with RoBERTa~\citep{liu2019roberta}, BART~\citep{Lewis_2020_bart}, 
Llama-2~\citep{touvron2023llama}, and Vistion Transformer~\citep{dosovitskiy2020_ViT}. 
We evaluate the performance of fine-turning strategies on natural language understanding (GLUE~\citep{Wang_2018_glue}, MMLU~\citep{hendrycks2020_mmlu}), natural language generation (XSum~\citep{Narayan_2018_xsum} and CNN/DailyMail~\citep{Chen_2016_cnndailymail}), and multi-domain image classification~\citep{gulrajani2020search}.

We implement all algorithms using PyTorch starting from the publicly-available Huggingface Transformers code base~\citep{wolf2019huggingface}. The conventional LoRA method applies a scaling coefficient $\alpha/r$ to $\Delta W$. Following LoRA~\citep{Hu2021LoRALA}, we fix $\alpha=2r$ to be twice the rank.  
Throughout our experiments, we use $\hat{A}$ to indicate matrix $A$ is being updated during fine-tuning and use subscripts \textit{\{rand, $0$, km\}} to indicate that the matrix is initialized as a random orthonormal matrix, zero matrix, and the random uniform initialization used in the original LoRA, respectively. 
Note that a properly normalized $d \times r$ random matrix with independent entries will have close to orthonormal columns when $d \gg r$ (see e.g. Theorem 4.6.1 of~\citet{vershynin2020high_hdp}), implying that the random orthonormal and random uniform initializations should be essentially equivalent.

We compare to the following methods:
\begin{enumerate}
    \item \textbf{Full fine-tuning (FT):} The most straightforward adaptation method, which initializes model parameters with the pre-trained weights and updates the whole model with gradient back-propagation.
    \item \textbf{Linear Probing (LP)~\citep{kumar2022finetuning}:}  A simple yet effective method that updates the last linear layer.
    \item  \textbf{IA$^3$~\citep{liu2022_ia3}:} Injects learned vectors in the attention and feedforward modules.  
    \item \textbf{LoRA:}~\citep{Hu2021LoRALA} Fine-tunes both $A$ and $B$ matrices of an additive $BA$ adaptation as introduced in previous sections, with a separate adaptation for each query/key/value parameter matrix.
    \item \textbf{AdaLora:}~\citep{zhang2023adalora} A variant of LoRA that adaptively changes the rank for each layer. 
\end{enumerate}

\subsection{Natural Language Understanding}
We use the General Language Understanding Evaluation~\citep[GLUE,][]{Wang_2018_glue} to evaluate the fine-tuning performance of different fine-tuning strategies. The GLUE benchmark contains a wide variety of tasks including question-answering, textual similarity, and sentiment analysis. We applied fine-tuning methods to the RoBERTa (large) model~\citep{liu2019roberta}, which has 355M parameters. To enable a fair comparison, we initialize the weights for all tasks with the original pretrained RoBERTa weights.

In Table \ref{tab:NLU_all_results} (see the appendix for an expanded table), we compare different freezing \& initialization strategies with LoRA and other baselines. 
We  \underline{underline} to indicate that performance is better than conventional LoRA also we use \textbf{bold} to denote the best performance when freezing one of the matrices. First, we can see a clear trend where solely updating the $B$ matrix outperforms just learning the $A$ matrix. In addition, when doubling the rank to match the trainable parameters,  $\mathbf{\hat{B}}_0 A_{orth}$ consistently outperforms conventional LoRA. This confirms our hypothesis in \S\ref{theoretical_discussion} that any loss in expressive power by not tuning $A$ can be made up for by the larger intrinsic rank of $B$ at no additional parameter cost. In fact, its performance statistically matches that of AdaLoRA, which uses over 3 times the parameters (incurring the associated memory and training costs). 


To assess the effects of different initialization methods for low-rank adaptation, we investigate different initialization methods thoroughly in Table \ref{tab:NLU_initialization}.  We can see that the best results always come from orthogonal initialization, which further supports our conclusions in \S\ref{sec:Analysis}.

\begin{table}[]
\begin{minipage}{0.48\columnwidth}
\centering
\caption{R-1/2/L (\%)
 on text summarization with BART-large on XSum and CNN/DailyMail. }
    \resizebox{0.99\columnwidth}{!}{%
  \begin{tabular}{l|r|cccc}
  \hline
  \toprule
Method & \# Param. & \multicolumn{2}{c}{} \\
         &  & XSum & CNN/DailyMail \\
  \midrule
  \midrule 
  
  
  $\hat{\mathbf{B}}_0 A_{rand, r=16}$ & {0.44} $\%$ & \textbf{42.91} / \textbf{19.61} / \textbf{34.64} &  \textbf{43.65} / \textbf{20.62} / \textbf{40.72}    \\

  $B_{rand} \hat{\mathbf{A}}_{0,r=16} $ & {0.44} $\%$ & 42.37 / 19.30 / 34.29 & 43.38 / 20.36 / 40.48    \\
 
\midrule
 
  $\hat{\mathbf{B}}_{0} \hat{\mathbf{A}}_{rand,r=8}$ & {0.44} $\%$ & 43.78 / \textbf{20.47} / \textbf{35.53} & 43.96 / 20.94 / 41.00   \\

  $\hat{\mathbf{B}}_{rand} \hat{\mathbf{A}}_{0,r=8}$  & {0.44} $\%$ & \textbf{43.80} / 20.39 / 35.48 & \textbf{44.07} / \textbf{21.08} / \textbf{41.19}  \\
  
  \bottomrule
  \end{tabular}
  }
  \label{tab_main:NLG}
\end{minipage}
\begin{minipage}{0.48\columnwidth}
\centering 
\caption{$5$-shot accuracy (\%) on the MMLU benchmark  
  }
 \resizebox{0.99\columnwidth}{!}{%
  \begin{tabular}{l|r|ccccc}
  \hline
  \toprule
Method & \# Param. & \multicolumn{5}{c}{5-shot} \\
         &  & Hums & STEM & Social & Other & Avg   \\
  \midrule
  \midrule
  Llama-2-7B & 100\% & 43.98 & 34.11 & 49.08 & 44.31 &  43.14   \\
  LoRA $_{r=32}$ & 0.24\% & 44.59 & 36.50 & 51.81 & \textbf{45.75} & {44.76}   \\
  
  \midrule
  \midrule
  
  $\hat{\mathbf{B}}_0 A_{rand,r=32}$ & 0.12\% & \textbf{44.17} & \textbf{36.00}  & \textbf{46.88} & 45.14 & \textbf{45.36} \\

  ${B}_{rand} \hat{\mathbf{A}}_{0,r=32}$ & 0.12\% & 44.36 & 35.93  & 51.46 & 46.85 & {44.51}  \\

  \midrule

  $\hat{\mathbf{B}}_0 A_{rand,r=64}$ & 0.12\% & \textbf{45.10} & \textbf{37.65}  & \textbf{55.08} & 51.08 & \textbf{46.46} \\
  
  \bottomrule
  \end{tabular}
  }
  \label{tab_main:mmlu}
\end{minipage} 
\end{table}

\begin{table*}[!t] 
\centering
\caption{DomainBed results (mean accuracy and standard deviation in $\%$). ID and OOD denote in-domain and out-of-domain test error, respectively. For OOD we report the average performance across different environments.
}
\resizebox{1.0\columnwidth}{!}{%
  \begin{tabular}{llccccccccc}
\toprule
Method & \# Param.  & \multicolumn{2}{c}{VLCS} & \multicolumn{2}{c}{PACS} & \multicolumn{2}{c}{OfficeHome} \\

 
 & &   (ID) & (OOD)  & (ID) & (OOD) & (ID) & (OOD)  \\
\midrule
\midrule

LoRA ${}_{r=8}$  & 0.46\% & 
 73.51\textsubscript{$\pm$0.62} & 56.43\textsubscript{$\pm$1.96} & 94.94\textsubscript{$\pm$0.56} & \textbf{75.58\textsubscript{$\pm$0.92}} & 
 78.54\textsubscript{$\pm$1.49} & 74.46\textsubscript{$\pm$0.40} 
  \\
  
  LP  & 0.00\% &  
 75.58\textsubscript{$\pm$1.66} & 71.70\textsubscript{$\pm$1.04} & 
 81.62\textsubscript{$\pm$0.34} & 61.73\textsubscript{$\pm$1.25} & 58.38\textsubscript{$\pm$0.76} & 68.59\textsubscript{$\pm$0.22}   \\

  Full Fine-tuning  & 100\% &  
 76.21\textsubscript{$\pm$1.95} & 64.87\textsubscript{$\pm$6.44} & 
 \textbf{98.15\textsubscript{$\pm$0.56}} & 74.90\textsubscript{$\pm$2.43} & \textbf{80.67\textsubscript{$\pm$1.22}} & 63.23\textsubscript{$\pm$0.64}   \\
 
 \midrule
\midrule

 $\hat{\textbf{B}} A_{rand,r=8}$   & 0.29\% &   77.40\textsubscript{$\pm$2.30} & \textbf{75.81\textsubscript{$\pm$1.65}}    
  & 92.45\textsubscript{$\pm$2.68} & 72.55\textsubscript{$\pm$1.03} & 
  77.66\textsubscript{$\pm$0.89} & 77.72\textsubscript{$\pm$0.32}    \\ 
  
 $\hat{\textbf{B}} A_{rand,r=16}$  & 0.46\% &  \textbf{79.10\textsubscript{$\pm$1.41}} & 75.40\textsubscript{$\pm$1.24} &   93.52\textsubscript{$\pm$0.20} & 73.76\textsubscript{$\pm$0.67} &  77.63\textsubscript{$\pm$0.84} & \textbf{77.85\textsubscript{$\pm$0.33}}   \\
 
 $B_{rand}\hat{\textbf{A}}_{r=8}$  & 0.29\% &  76.71\textsubscript{$\pm$0.93} & 72.50\textsubscript{$\pm$0.89}
 & 92.02\textsubscript{$\pm$1.07} & 66.25\textsubscript{$\pm$0.80} & 72.36\textsubscript{$\pm$0.69} & 73.66\textsubscript{$\pm$0.35}   \\

\bottomrule
\end{tabular}}
\label{tab:domainbed_new}
\end{table*}

  

\subsection{Natural Language Generation}
To investigate the asymmetry of low-rank fine-tuning in natural language generation (NLG), we fine-tune a BART-large model~\citep{Lewis_2020_bart} and evaluate model performance on the XSum~\citep{Narayan_2018_xsum} and CNN/DailyMail~\citep{Chen_2016_cnndailymail} datasets. Following \citet{zhang2023adalora}, we apply low-rank adaptation to every {query/key/value} matrix and report ROUGE 1/2/L scores (R-1/2/L,~\citep{lin2004rouge}). We fine-tune models for $15$ epochs. We select the beam length as $8$ and batch size as {$48$} for XSum, and the beam length as $4$, batch size as {$48$} for CNN/DailyMail. More details of the configurations are in the Appendix \ref{appendix:text-summarization}. 

The results are summarized in Table \ref{tab_main:NLG}. In the first two rows, we observe the asymmetry between the factors since freezing $A$ and only updating $B$ always outperforms only updating $A$. The last two rows show the results of tuning both matrices with different initializations, showing that the asymmetry is not explained by the initialization strategy.

\subsection{Massive Multitask Language Understanding}
We fine-tune the pretrained Llama-2-7B model~\citep{touvron2023llama} using instruction tuning on the Alpaca dataset~\citep{wang2023far_alpaca}. We assess the asymmetry on the MMLU benchmark~\citep{hendrycks2020_mmlu}, which consists of 57 distinct language tasks. As shown in Table \ref{tab_main:mmlu}, the asymmetry also exists in larger language models, and updating $B$ consistently outperforms updating $A$. Moreover, it also outperforms standard LoRA except for ``Other'' where it matches the performance, reflecting the benefits of being able to increase $r$ without tuning more parameters.



\subsection{Vision Transformers and Generalization}

We next measure generalization, motivated by the theory in \S\ref{Advantages of tuning $B$ only over $BA$ together}. In particular, we work with ViTs in image classification tasks using the Domainbed testbed for domain generalization~\cite{gulrajani2020search}. Domainbed contains several datasets, each composed of multiple environments (or domains). Classes in each environment tend to be similar at a high level but differ in terms of style. We fine-tune a pre-trained ViT, originally trained on ImageNet, on the LabelMe, Cartoon, and Clipart environments within the VLCS, PACS, and Office-Home datasets, respectively. We employ different benchmark fine-tuning methods such as full fine-tuning, linear probing, and LoRA, and compare their performance to freezing either $A$ or $B$ in in-domain and out-of-domain generalization. We adhere to the original $80\%$ training and $20\%$ testing splits. 

Results are presented in Table~\ref{tab:domainbed_new} (see Appendix~\ref{Additional Domainbed results} for extended version). In line with our expectations, randomly initializing and freezing matrix $A$ while only updating matrix $B$ generally results in better out-of-domain test accuracy. We report additional generalization results in Appendix~\ref{Additional Domainbed results}, in which we compare the train set and test set accuracy of the different approaches. We consistently find that fine-tuning a single matrix leads to smaller gaps between these two quantities compared to LoRA, paralleling the corresponding reduction in the generalization bounds of \S\ref{Advantages of tuning $B$ only over $BA$ together}.



\section{Conclusion}

In this paper, we formally identify and investigate asymmetry in the roles of low-rank adapter matrices in LoRA fine-tuning. The $A$ matrices extract features from the input, while the $B$ matrices project these features towards the desired objective. We illustrate differences between the two matrices from both theoretical and empirical perspectives. Our theoretical analysis explains the asymmetry in the fine-tuning of large models and suggests that freezing $A$ as a random orthogonal matrix can improve generalization, a claim we corroborate with experiments across multiple models and datasets. Our work serves as an initial step to unveil the mechanisms of fine-tuning large models, and it provides an understanding that can benefit future research directions, promoting efficiency and interpretability.





\bibliography{example_paper}
\bibliographystyle{icml2024}

\newpage
\appendix
\section{Similarity Metric in Figure \ref{fig1}}
\label{app:fig1}
To measure the similarity of learned $A$ and $B$ matrices we adopted a measure that accounts for the invariance of LoRA fine-tuning. Let $\Delta W = BA$ denote the learned LoRA adapter. Since $BA = BCC^{-1}A$ for any invertible matrix $C \in \mathbb{R}^{r \times r}$, we can define $\tilde{B} = BC$ and $\tilde{A} = C^{-1}A$ resulting in the same LoRA adapter $\Delta W = \tilde{B} \tilde{A}$. Thus, to measure the similarity of LoRA matrices we need a metric that is invariant to invertible linear transformations, i.e., dissimilarity$(B,BC) = 0$ for any invertible $C$. In our experiment, we used Canonical Correlation Analysis goodness of fit \citep{ramsay1984matrix}, similar to prior work comparing neural network representations \citep{kornblith2019similarity}. The key idea is to compare orthonormal bases of the matrices, thus making this similarity metric invariant to invertible linear transformations. 

More specifically, given two matrices $X \in \mathbb{R}^{n \times r_1}$ and $Y \in \mathbb{R}^{n \times r_2}$, the similarity is computed as follows: $\|U_Y^\top U_X \|^2_F / \min \{r_1, r_2\}$, where $U_X / U_Y$ is the orthonormal bases for the columns of $X / Y$. Following a similar method as in \citet{Hu2021LoRALA}, for $A$ we perform SVD and use the right-singular unitary matrices as the bases, and use left-singular unitary matrices for $B$.

\subsection{Reversed Initialization}

The initialization of adapter matrices can play an important role in LoRA fine-tuning. To further investigate the effect of initialization on asymmetry, we reverse the initialization compared to conventional LoRA, where $A$ is initialized to zero and $B$ is initialized with random uniform distributions. Overall, we observe that the trend of differences also reverses, which is expected given the significant role of initialization in training deep learning models. 

\begin{figure}[htbp!]
\centering
    \subfigure[Random initialization, same task]{\includegraphics[width=0.31\textwidth]{figures/sametask_diffseed_cca_mean.png} \label{fig1_apdx:same_task}
    } 
    \subfigure[Fixed initialization, different tasks
    ]{\includegraphics[width=0.31\textwidth]{figures/difftask_sameseed_cca_mean.png}
    \label{fig1_apdx:diff_task}} 
    \subfigure[Random initialization, different tasks]{\includegraphics[width=0.31\textwidth]{figures/difftask_diffseed_cca_mean.png} \label{fig1_apdx:sametask_diffseed}
    } 
    
    \centering
    \subfigure[Random initialization, same task]{\includegraphics[width=0.31\textwidth]{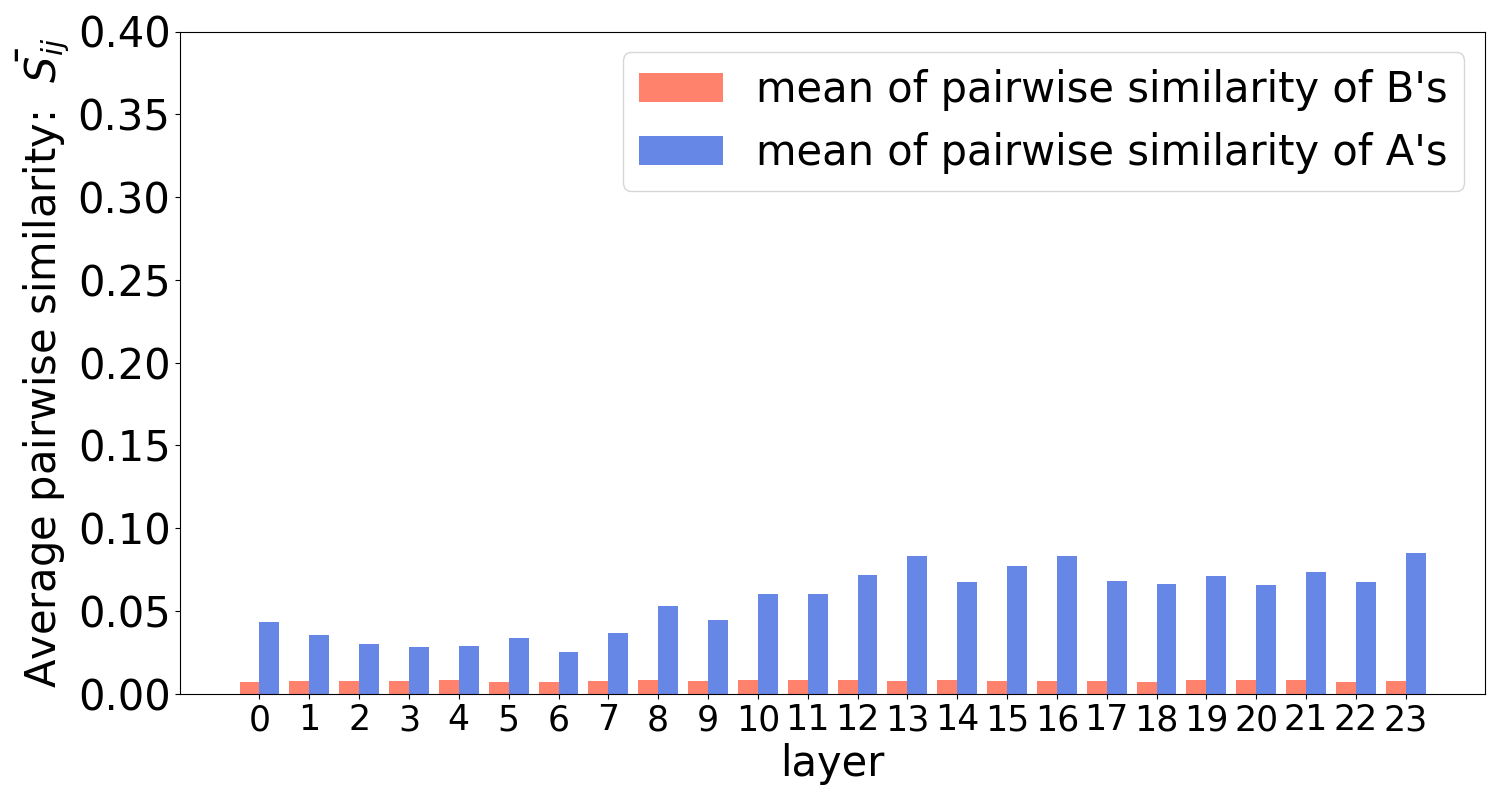} \label{fig1_apdx:reverse_same_task}
    } 
    \subfigure[Fixed initialization, different tasks
    ]{\includegraphics[width=0.31\textwidth]{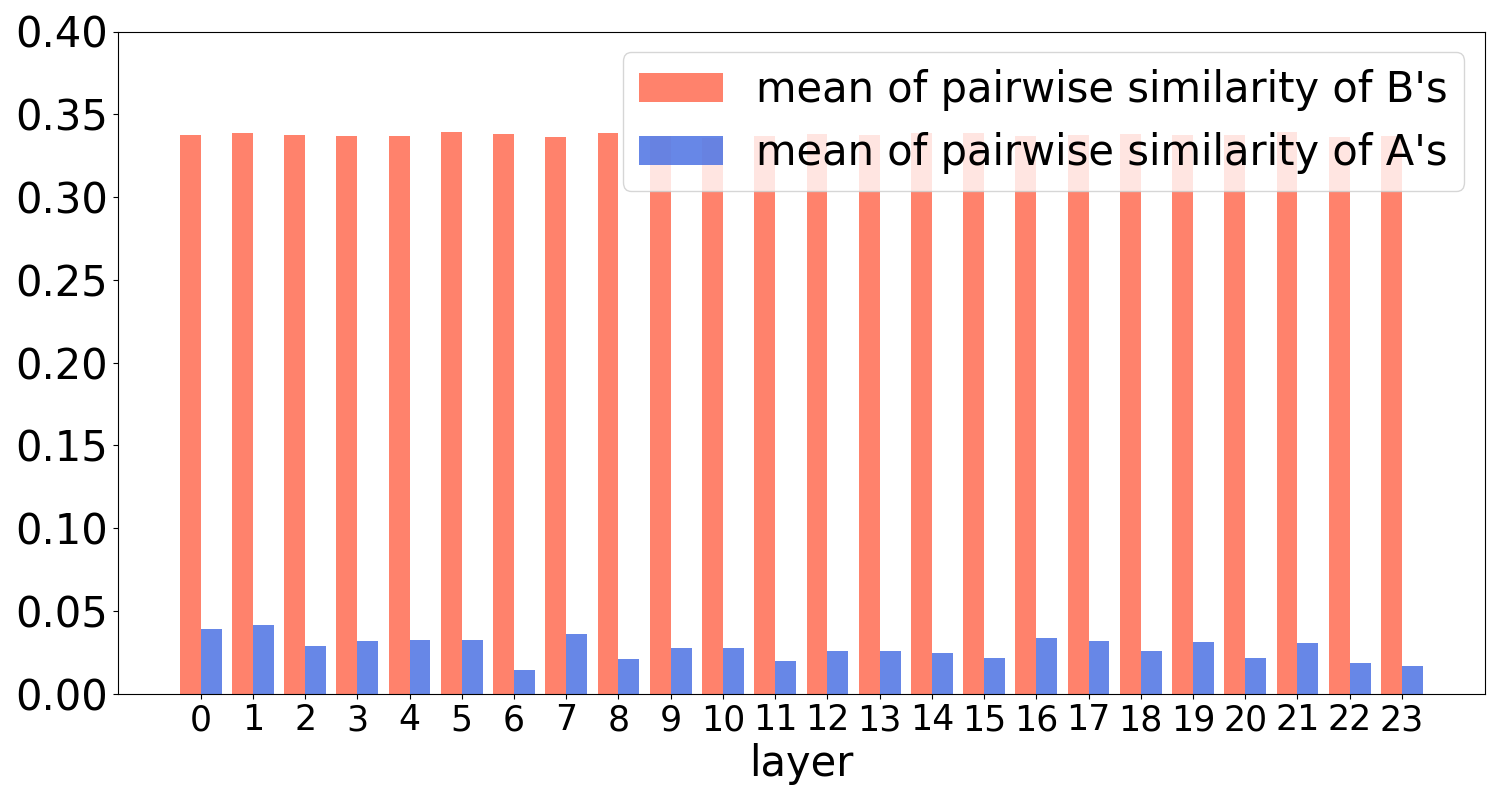}
    \label{fig1_apdx:reverse_diff_task}} 
    \subfigure[Random initialization, different tasks]{\includegraphics[width=0.31\textwidth]{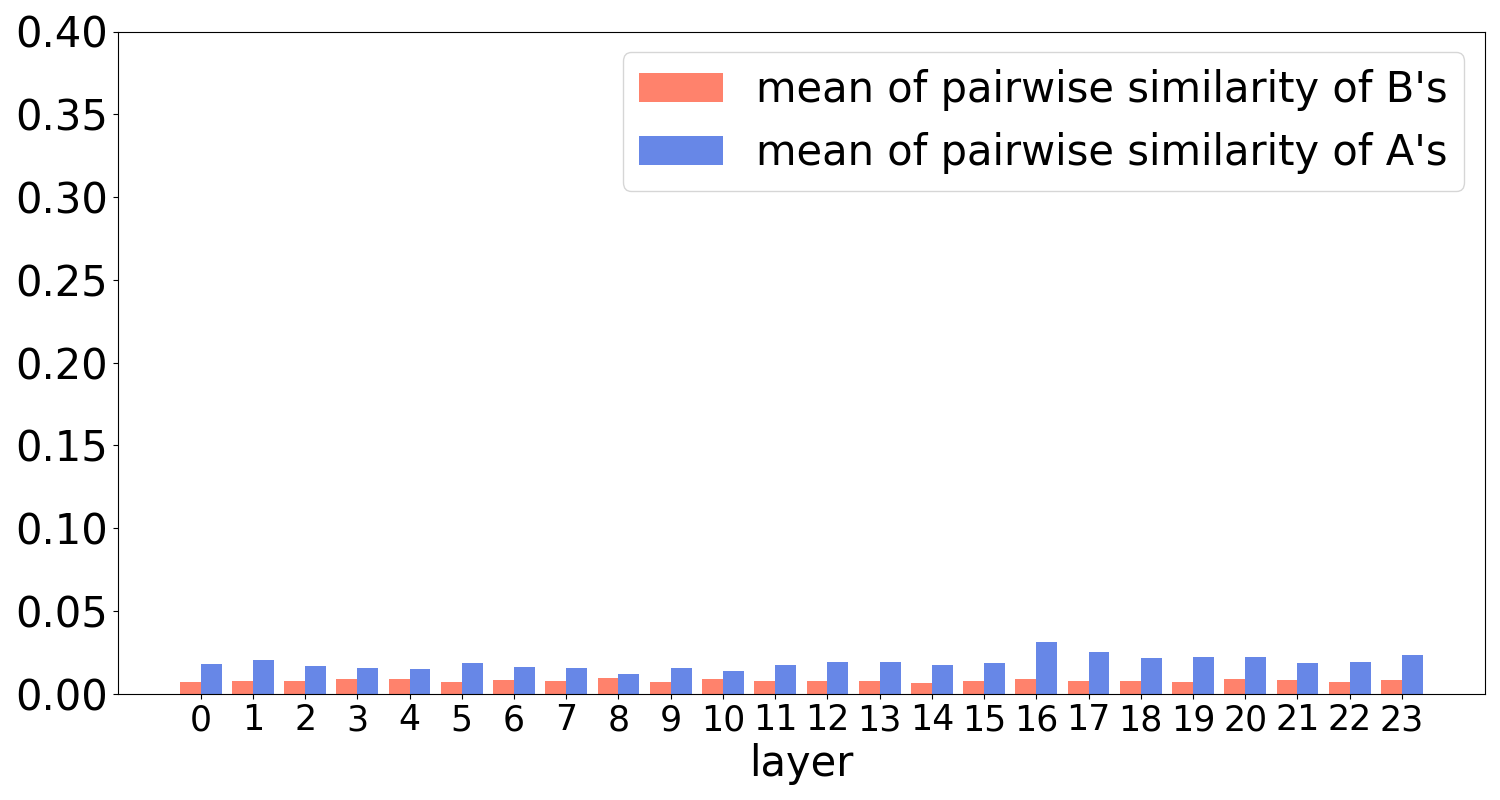} \label{fig1_apdx:reverse)sametask_diffseed}
    } 
    \caption{Similarity of learned LoRA matrices $A$ \& $B$ across layers of a RoBERTa model fine-tuned with different initialization and data settings. We compare the results from both conventional LoRA initialization (In Figure (a), (b), and (c), $A$ is initialized as random uniform $B$ is initialized as zero) and a reversed initialization (In Figure (d), (e), and (f), $A$ is initialized as zero $B$ is initialized as random uniform. 
}
    \label{fig1_reveresed}
\end{figure}

When comparing the similarities of different initialization strategies, we can still draw the same conclusion about the importance of the $B$ matrix. For example, compared with Figure \ref{fig1_apdx:same_task}, the $A$ matrices in Figure \ref{fig1_apdx:reverse_same_task} have a smaller similarity in average. Such difference can also be observed when comparing Figure \ref{fig1_apdx:diff_task} and \ref{fig1_apdx:reverse_diff_task}.

\section{Asymmetry Proofs for Multivariate Least Squares}
\label{app:LS}

\subsection{Proof of Lemma \ref{thm:B_L2}}
Consider freezing $B = U$ where $U$ is orthogonal ($U^\top  U = I_r$) and fine-tuning $A$. The objective becomes
\begin{align*}
&A^\ast = \arg\min_{A} \mathcal{L}(A, U) \\
&= \arg\min_{A}\mathbb{E}_{(Y_{targ}, X_{targ})} \left\| Y_{targ} - (W_0 + U A) X_{targ} - b \right\|_2^2\\
&= \arg\min_{A}\mathbb{E} \left\| (W_{targ} X_{targ} - W_0 X_{targ}) - U A X_{targ} \right\|_2^2\\
&= \arg\min_{A}\mathbb{E} \left\| U^\top  ((W_{targ} - W_0) X_{targ} + n) - A X_{targ} \right\|_2^2\\
&= U^\top  \Delta.
\end{align*}

Interestingly, note that this solution $A^\ast$ does not depend on the distribution of $X_{targ}$, it is simply the projection of the difference between the pretrained $W_0$ and the target $W_{targ}$. This is because, intuitively, freezing $B$ is projecting down the \emph{outputs} into $r$ dimensional space, and then optimizing $A$ to match these projected outputs. It can be shown that the expected squared prediction error is
\begin{align*}
    \mathcal{L}(A^\ast,U) = d_{out} \sigma^2 + \mathrm{Tr}[\Delta \Sigma \Delta^\top ] - \mathrm{Tr}[U^\top  \Delta \Sigma \Delta^\top  U],
\end{align*}
where $\Sigma = \mathrm{Cov}[ X_{targ} ]$.

\subsection{Proof of Lemma \ref{thm:A_L2}} Consider freezing $A = Q$ where $Q$ is orthogonal ($Q Q^\top  = I_r$) and fine-tuning $B$. The objective becomes
\begin{align*}
&B^\ast = \arg\min_{B} \mathcal{L}(Q, B) \\
&= \arg\min_{B}\mathbb{E}_{(Y_{targ}, X_{targ})} \left\| Y_{targ} - (W_0 + B Q) X_{targ} \right\|_2^2\\
&= \arg\min_{B}\mathbb{E} \left\| (Y_{targ} - W_0 X_{targ}) - B (Q X_{targ}) \right\|_2^2,
\end{align*}
which is simply an ordinary least squares regression problem mapping $Q X_{targ}$ to $(Y_{targ} - W_0 X_{targ})$. The solution is known to be
\[
B^\ast = \Delta \Sigma Q^\top  (Q \Sigma Q^\top )^{-1}
\]
yielding an expected squared prediction error of
\begin{align*}
 \mathcal{L}(Q,B^{\ast})  &= d_{out} \sigma^2 + \mathrm{Tr}[\Delta \Sigma \Delta^\top ] - \mathrm{Tr} [Q \Sigma \Delta^\top  \Delta \Sigma Q^\top  (Q \Sigma Q^\top )^{-1}].
\end{align*}
Note that the solution is now clearly dependent on the distribution of $X_{targ}$, and the first two terms of the squared prediction error are the same but the third term is different. 

\subsection{Proof of Theorem \ref{thm:ABcomp}}
The third term in the expression for freezing $A$ is
\begin{align*}
III_A = &\mathrm{Tr} [Q \Sigma \Delta^\top  \Delta \Sigma Q^\top  (Q \Sigma Q^\top )^{-1}] \\
&\geq \mathrm{Tr} [Q \Sigma \Delta^\top  \Delta Q^\top  Q \Sigma Q^\top  (Q \Sigma Q^\top )^{-1}]\\
&= \mathrm{Tr} [Q \Sigma \Delta^\top  \Delta Q^\top ],
\end{align*}
where the inequality follows by Von Neumann's trace inequality and the fact that the product of two positive semidefinite matrices has nonnegative real eigenvalues.
Compare to the third term in the expression for freezing $B$:
\[
III_{B} = \mathrm{Tr}[U^\top  \Delta \Sigma \Delta^\top  U].
\]
Recall that $U, Q$ are drawn uniformly at random from their respective Stiefel manifolds.
Then
\[
\mathbb{E} [III_B] \rightarrow \frac{r}{d} \mathrm{Tr}[\Delta \Sigma \Delta^\top ]
\]
and we have
\begin{align*}
&\mathbb{E} [III_A] \geq \mathbb{E}[\mathrm{Tr} [Q \Sigma \Delta^\top  \Delta Q^\top ]]  \rightarrow \frac{r}{d} \mathrm{Tr}[ \Sigma \Delta^\top  \Delta] = \frac{r}{d} \mathrm{Tr}[\Delta \Sigma \Delta^\top  ] \rightarrow \mathbb{E} [III_B].
\end{align*}
Hence $\lim_{d/r\rightarrow \infty} \mathbb{E} [III_A] \geq \lim_{d/r\rightarrow \infty} \mathbb{E} [III_B]$, implying that freezing $A$ to a random orthogonal matrix achieves lower mean squared error loss than freezing $B$.

\section{Proof of Lemma \ref{lemm:genbd}: Generalization Bounds}
\label{app:IT}
We use the following bound on the generalization error is from \cite{xu2017}, specialized to our setting and notation.
\begin{theorem}[specialized from \cite{xu2017}]
Denote by $\mathcal{A}$ a LoRA-based fine-tuning algorithm, which outputs $\Delta \mathbf{W}$ given $S_n$.  Assume that 
$\ell^{\mathbf{W}, \mathbf{b}}(\Delta \mathbf{W}, \widetilde{Z})$ is $\sigma$-sub-Gaussian under $(\Delta \mathbf{W}, \widetilde{Z}) \sim P_{\Delta \mathbf{W} | \mathbf{W}, \mathbf{b}} \times \mu$. Then,
\begin{equation}
    | \generror{\mu}{\mathcal{A}} | \!\leq\! \sqrt{\frac{{2\sigma^2}}{n}   \mathsf{I}(\Delta \mathbf{W};S_n | \mathcal{A},\mathbf{W}}).
\end{equation}
\end{theorem}
We consider the case of tuning $B$ only first. Applying the above theorem, note that here
\begin{align*}
\mathsf{I}(\Delta \mathbf{W};S_n | \mathcal{A}_{B},\mathbf{W}) &= \mathsf{I}(\{B_i Q_i\}_{i\in\mathcal{I}};S_n | \mathcal{A}_{B},\mathbf{W}) \\&= \mathsf{I}(\{B_i\}_{i\in\mathcal{I}};S_n | \mathcal{A}_{B},\mathbf{W}),
\end{align*}
where we have used the data processing inequality (DPI), noting that the $Q_i$ are here considered orthogonal fixed constant matrices as they are not trained, hence the mapping from $B_i$ to $B_i Q_i$ is invertible.

We can now bound this expression as
\begin{align*}
   \mathsf{I}(\{B_i\}_{i\in\mathcal{I}};S_n | \mathcal{A}_{B},\mathbf{W}) &\leq H(\{B_i\}_{i\in\mathcal{I}})\\
&\leq q r\sum_{i\in\mathcal{I}} d_{out}^{(i)},
\end{align*}
where we have noted that mutual information is upper bounded by discrete entropy, and entropy in turn is upper bounded by the uniform distribution over its possible support set ($q$ bits in each of $r\sum_{i\in \mathcal{I}} d_{out}^{(i)}$ dimensions). The bounds for the other two algorithms are similar.






\section{Text Generation Training Details}
\label{appendix:text-summarization}

The configuration of our experiments on text generation is listed in Table \ref{tab:NLG-details}.

\begin{table}[h!]
  \centering
  \addtolength{\tabcolsep}{-4pt}
  
  \caption{Hyper-parameter setup for summarization tasks. 
  }
  \label{tab:NLG-details}
  \begin{tabular}{l|ccccccc}
  \hline
  \toprule
Dataset & learning rate & batch size & \# epochs & $\gamma$ & $t_i$ & $\Delta_{T}$ & $t_f$ \\
  \midrule
  \midrule 
  
  
  \textbf{XSum} & $5\times10^{-4}$ & $48$ & $25$ & $0.1$ & $6000$ & $100$ & $50000$  \\
  \textbf{CNN/DailyMail} & $5\times10^{-4}$ & $48$ & $15$ & $0.1$ & $5000$ & $100$ & $85000$  \\

  \bottomrule
\end{tabular}
\end{table}

\clearpage
\section{Additional Language Results}
See Table \ref{tab:NLU_all_results_app} for additional results.
\begin{table*}[htbp!]
  \centering
  \addtolength{\tabcolsep}{-4pt}
  
  \caption{Different adaptation methods on the GLUE benchmark. We report the overall (matched and mismatched) accuracy for MNLI, Matthew's correlation coefficient for CoLA, Pearson correlation for STS-B, and accuracy for other tasks. Higher is better for all metrics. 
  }
  \label{tab:NLU_all_results_app}
  \begin{tabular}{l|r|ccccccccc}
  \hline
  \toprule
  Model \& Method & \# Trainable & \multicolumn{8}{c}{} \\
         & Parameters & MNLI & SST-2 & MRPC & CoLA & QNLI & RTE & STS-B & Avg. \\
         
  
  \midrule 
  LoRA $(r=8)$ & 0.8M & 90.3\textsubscript{$\pm$.07} & 95.6\textsubscript{$\pm$0.36} & {90.3}\textsubscript{$\pm$0.85} & 64.4\textsubscript{$\pm$1.8} & {94.0}\textsubscript{$\pm$0.29} &   {84.1}\textsubscript{$\pm$0.96}  & {91.5}\textsubscript{$\pm$0.16} & 87.2 \\
  
  AdaLoRA & 2.5\% & 90.4\textsubscript{$\pm$.37}  & 95.9\textsubscript{$\pm$.13} & 90.1\textsubscript{$\pm$.54} & {67.5}\textsubscript{$\pm$1.3} & {94.7}\textsubscript{$\pm$.22} & {85.4}\textsubscript{$\pm$.20} &  {91.3}\textsubscript{$\pm$1.0}  & 87.9 \\
  
  $(\text{IA})^3$ & 0.7\% & 90.0\textsubscript{$\pm$.21} & 95.4\textsubscript{$\pm$.17} & 83.7\textsubscript{$\pm$.13} & {57.6}\textsubscript{$\pm$.67} & {93.7}\textsubscript{$\pm$.07} & {70.3}\textsubscript{$\pm$1.5} & {87.0}\textsubscript{$\pm$0.4} & 82.5\\
  
  \midrule
  
  $\hat{B}_0 A_{V}$ $(r=8)$ & 0.3M & 90.1\textsubscript{$\pm$.09}  & 95.5\textsubscript{$\pm$.01}  & 90.8\textsubscript{$\pm$.24} & {63.8}\textsubscript{$\pm$4.2} & {94.2\textsubscript{$\pm$.11}} &   83.3\textsubscript{$\pm$1.7}  & 91.3\textsubscript{$\pm$.24}  & 87.0 \\
  
  $\hat{B}_0 A_{rand}$ $(r=8)$ & 0.3M & 90.1\textsubscript{$\pm$.19} &  95.8\textsubscript{$\pm$.29} & 89.7\textsubscript{$\pm$.13} & {67.5}\textsubscript{$\pm$1.2} & 94.0\textsubscript{$\pm$.27} &  82.8\textsubscript{$\pm$1.5}   & {91.9}\textsubscript{$\pm$.26} & 87.4  \\
  
  $\hat{B}_0 A_{km}$ $(r=8)$  & 0.3M & 90.1\textsubscript{$\pm$.17}  & 95.6\textsubscript{$\pm$.17} & 90.6\textsubscript{$\pm$.32} & {67.3}\textsubscript{$\pm$2.3}  & 93.4\textsubscript{$\pm$.61}  &    82.4\textsubscript{$\pm$1.4}  &  91.2\textsubscript{$\pm$.29} & 87.2 \\
  
  ${B}_U \hat{A}_0$ $(r=8)$& 0.3M & 89.3\textsubscript{$\pm$.18}  & 95.4\textsubscript{$\pm$0.13} & 88.8\textsubscript{$\pm$0.70}  & {59.1}\textsubscript{$\pm$0.48} & {93.8}\textsubscript{$\pm$0.15} &   {77.5}\textsubscript{$\pm$2.7} & 90.7\textsubscript{$\pm$.27} & 94.9 \\
  
  ${B}_{rand} \hat{A}_0$ $(r=8)$ & 0.3M & 90.3\textsubscript{$\pm$.18} & 95.5\textsubscript{$\pm$.66} & 89.3\textsubscript{$\pm$.09} &   {58.7}\textsubscript{$\pm$2.5} & {93.8}\textsubscript{$\pm$.21} &   {77.1}\textsubscript{$\pm$1.3} & 90.7\textsubscript{$\pm$.31} &  85.1 \\
  
  ${B}_{km} \hat{A}_0$ $(r=8)$  & 0.3M & 34.5\textsubscript{$\pm$1.6} & 95.2\textsubscript{$\pm$.34}  & 89.3\textsubscript{$\pm$.11} & 0.0\textsubscript{$\pm$0.0} & {93.0}\textsubscript{$\pm$.38} &   {47.3}\textsubscript{$\pm$.0} & 91.2\textsubscript{$\pm$.24} & 64.4 \\
  
  \midrule
  
  $\hat{B}_0 A_{V}$ $(r=16)$ & 0.8M & 90.2\textsubscript{$\pm$.17} & 95.8\textsubscript{$\pm$.20} & 90.1\textsubscript{$\pm$.56} & {67.8}\textsubscript{$\pm$.49} & 94.5\textsubscript{$\pm$.07} &   {82.8}\textsubscript{$\pm$.42} & 91.6\textsubscript{$\pm$.21} & 87.5 \\
  
  $\hat{B}_0 A_{rand}$ $(r=16)$ & 0.8M & 90.1\textsubscript{$\pm$.20} & {96.1\textsubscript{$\pm$.18}} & {90.7}\textsubscript{$\pm$.90} & {66.1}\textsubscript{$\pm$2.6} & {94.4\textsubscript{$\pm$.10}}  &   {84.1}\textsubscript{$\pm$.96}  & 91.2\textsubscript{$\pm$.42}  & 87.5 \\
  
   $\hat{B}_0 A_{km}$ $(r=16)$ & 0.8M & 90.3\textsubscript{$\pm$.06} & {95.6\textsubscript{$\pm$.01}}  & 91.1\textsubscript{$\pm$.32} & 65.2\textsubscript{$\pm$2.1} &  94.5\textsubscript{$\pm$.02} &   {81.7}\textsubscript{$\pm$1.8} &  91.2\textsubscript{$\pm$.39} & 87.1 \\

  ${B}_U \hat{A}_0$ $(r=16)$& 0.8M & 90.3\textsubscript{$\pm$.07}  & 95.4\textsubscript{$\pm$.57} & 90.4\textsubscript{$\pm$1.1} & 60.7\textsubscript{$\pm$.14} & 94.1\textsubscript{$\pm$.30} &   {80.1}\textsubscript{$\pm$1.2} & 90.8\textsubscript{$\pm$.29} & 86.0 \\
  
  ${B}_{rand} \hat{A}_0$ $(r=16)$ & 0.8M & 
  89.9\textsubscript{$\pm$.19}& 95.6\textsubscript{$\pm$.64} & 90.2\textsubscript{$\pm$0.23} &  60.3\textsubscript{$\pm$3.3} & 93.9\textsubscript{$\pm$0.25} &   {80.4}\textsubscript{$\pm$0.21} & 90.9\textsubscript{$\pm$0.13} & 85.9  \\
  
  ${B}_{km} \hat{A}_0$ $(r=16)$  & 0.8M & 89.2\textsubscript{$\pm$.03} & 95.2\textsubscript{$\pm$.29}  & 90.6\textsubscript{$\pm$0.65} & 40.4\textsubscript{$\pm$35.} & 93.1\textsubscript{$\pm$0.23}  &   {70.3}\textsubscript{$\pm$0.19} & 91.4\textsubscript{$\pm$0.26} & 81.5 \\
  \midrule 

  $\hat{B}_0 \hat{A}_{V}$ $(r=8)$ & 0.8M & 90.4\textsubscript{$\pm$.11} & 95.9\textsubscript{$\pm$0.18}  & 90.7\textsubscript{$\pm$0.84} & 64.0\textsubscript{$\pm$0.50} & 94.4\textsubscript{$\pm$0.16} &   {84.1}\textsubscript{$\pm$0.15} & 91.8\textsubscript{$\pm$00.15} & 87.3 \\
  
  $\hat{B}_0 \hat{A}_{rand}$ $(r=8)$ & 0.8M & 90.4\textsubscript{$\pm$.15} & 96.0\textsubscript{$\pm$.63}  & 91.5\textsubscript{$\pm$1.1} & 64.1\textsubscript{$\pm$0.67} & 94.5\textsubscript{$\pm$0.11} &   {85.6}\textsubscript{$\pm$0.96} & 92.0\textsubscript{$\pm$0.31} & 87.7 \\
  
  $\hat{B}_0 \hat{A}_{km}$ $(r=8)$  & 0.8M & 90.3\textsubscript{$\pm$.07} & 95.6\textsubscript{$\pm$0.36} & 90.3\textsubscript{$\pm$0.85} & 64.4\textsubscript{$\pm$1.8} & 94.0\textsubscript{$\pm$0.29} &   {84.1}\textsubscript{$\pm$0.96} & 91.5\textsubscript{$\pm$0.16} & 87.2 \\
  
  $\hat{B}_U \hat{A}_0$ $(r=8)$& 0.8M & 
  90.3\textsubscript{$\pm$.11}& 96.1\textsubscript{$\pm$.18} & 91.7\textsubscript{$\pm$0.33} & 64.9\textsubscript{$\pm$1.5}  & 94.7\textsubscript{$\pm$0.33}  &   {84.8}\textsubscript{$\pm$0.96} & 91.9\textsubscript{$\pm$0.19} & 87.8 \\
  
  $\hat{B}_{rand} \hat{A}_0$ $(r=8)$ & 0.8M & 90.3\textsubscript{$\pm$.27} & 96.0\textsubscript{$\pm$.26} & 90.8\textsubscript{$\pm$0.51} & 66.0\textsubscript{$\pm$1.01} & 94.5\textsubscript{$\pm$0.38} &   {83.6}\textsubscript{$\pm$1.5} & 92.0\textsubscript{$\pm$0.18}  & 87.6  \\
  
  $\hat{B}_{km} \hat{A}_0$ $(r=8)$  & 0.8M & 35.5\textsubscript{$\pm$1.6} & 95.6\textsubscript{$\pm$.65} & 90.0\textsubscript{$\pm$0.46} & 21.3\textsubscript{$\pm$36.} & 93.8\textsubscript{$\pm$0.01} &   {57.4}\textsubscript{$\pm$0.17} & 91.6\textsubscript{$\pm$0.43} & 69.3 \\
  
  \bottomrule
  \end{tabular}
\end{table*}

\section{Additional Vision Transformers and Generalization Results}
\label{Additional Domainbed results}

Table~\ref{tab:domainbed_original} displays a more fine-grained version of Table~\ref{tab:domainbed_new} in the main text, and presents results for each out-of-distribution environment independently, in which it is easier to appreciate the benefits of only updating $B$ in terms of out-of-domain performance. Additional results for TerraIncognita, as well as generalization results, can be found in Table~\ref{tab:domainbed_extra} and Table~\ref{tab:domainbed_generalization}, respectively. TerraIncognita seems to be a particularly challenging dataset to which low-rank adapters struggle to fit; the most effective method, in this case, appears to be full fine-tuning. In terms of generalization, we can observe that fine-tuning only a single adapter matrix generally results in a lower difference between training set and test set accuracy compared to standard LoRA for all datasets.

\begin{table*}[htbp!] 
\centering
\caption{DomainBed results (mean accuracy and standard deviation in $\%$). ID and OOD denote in-domain and out-of-domain generalization, respectively. 
}
\resizebox{\columnwidth}{!}{%
  \begin{tabular}{llccccccccccccc}
\hline
Method & \# Trainable Parameters   & \multicolumn{4}{c}{VLCS} & \multicolumn{4}{c}{PACS} & \multicolumn{4}{c}{OfficeHome} \\
 & (\% full ViT params) & Caltech101  & LabelMe  & SUN09  & VOC2007  & Art  & Cartoon  & Photo  & Sketch  & Art  & Clipart  & Product  & Photo  \\
 & & (OOD) & (ID) & (OOD) & (OOD) & (OOD) & (ID) & (OOD) & (OOD) & (OOD) & (ID) & (OOD) & (OOD) \\
\hline
 $\hat{B} A_{rand}$ $(r=8)$  & 0.16M-0.2M (0.18-0.29\%) & \textbf{93.19\textsubscript{$\pm$2.27}} & 77.40\textsubscript{$\pm$2.30} & \textbf{61.52\textsubscript{$\pm$1.50}} & 72.72\textsubscript{$\pm$1.18} & 
 81.22\textsubscript{$\pm$1.40} & 92.45\textsubscript{$\pm$2.68} & 96.07\textsubscript{$\pm$0.86} & 40.37\textsubscript{$\pm$0.83}
 & 73.59\textsubscript{$\pm$0.59} & 77.66\textsubscript{$\pm$0.89} & 78.02\textsubscript{$\pm$0.14} & 81.55\textsubscript{$\pm$0.24}   \\ 
 $\hat{B} A_{rand}$ $(r=16)$ & 0.3M-0.4M (0.36-0.46\%) & 91.57\textsubscript{$\pm$0.81} & \textbf{79.10\textsubscript{$\pm$1.41}} & 60.97\textsubscript{$\pm$2.44} & \textbf{73.66\textsubscript{$\pm$0.46}} & \textbf{84.36\textsubscript{$\pm$0.54}} & 93.52\textsubscript{$\pm$0.20} &\textbf{97.07\textsubscript{$\pm$0.47}}&39.87\textsubscript{$\pm$0.99}
 
 & \textbf{73.64\textsubscript{$\pm$0.40}}&77.63\textsubscript{$\pm$0.84}&\textbf{78.07\textsubscript{$\pm$0.22}}&\textbf{81.85\textsubscript{$\pm$0.36}} \\
 $B_{rand}\hat{A}$ $(r=8)$ & 0.16M-0.2M (0.18-0.29\%) & 87.18\textsubscript{$\pm$0.77}&76.71\textsubscript{$\pm$0.93}&59.89\textsubscript{$\pm$1.79}&70.44\textsubscript{$\pm$0.10} & 
 77.05\textsubscript{$\pm$0.74}&92.02\textsubscript{$\pm$1.07}&92.06\textsubscript{$\pm$0.34}&29.65\textsubscript{$\pm$1.31}
 & 68.36\textsubscript{$\pm$0.28}&72.36\textsubscript{$\pm$0.69}&74.00\textsubscript{$\pm$0.31}&78.63\textsubscript{$\pm$0.45} \\
$B_{rand}\hat{A}$ $(r=16)$  & 0.3M-0.4M (0.36-0.46\%) & 89.28\textsubscript{$\pm$2.51}&78.03\textsubscript{$\pm$1.23}&60.44\textsubscript{$\pm$1.84}&70.81\textsubscript{$\pm$0.36} & 
81.43\textsubscript{$\pm$0.92}&93.87\textsubscript{$\pm$0.73}&95.63\textsubscript{$\pm$0.13}&35.02\textsubscript{$\pm$0.86}
& 71.64\textsubscript{$\pm$0.24}&73.77\textsubscript{$\pm$1.13}&75.46\textsubscript{$\pm$0.25}&80.31\textsubscript{$\pm$0.39}\\ \hline
  LoRA $(r=8)$  & 0.3M-0.4M (0.35-0.46\%) & 
    44.59\textsubscript{$\pm$1.96}&73.51\textsubscript{$\pm$0.62}&60.44\textsubscript{$\pm$2.86}&64.26\textsubscript{$\pm$1.07}
  & 81.41\textsubscript{$\pm$0.70}&\textbf{94.94\textsubscript{$\pm$0.56}}&95.43\textsubscript{$\pm$0.54}&49.90\textsubscript{$\pm$1.51}
  & 70.44\textsubscript{$\pm$0.46}&78.54\textsubscript{$\pm$1.49}&73.99\textsubscript{$\pm$0.64}&78.95\textsubscript{$\pm$0.10}

  \\
  Linear Probing  & 0.004M (0.00\%) &  
  90.65\textsubscript{$\pm$2.51}&75.58\textsubscript{$\pm$1.66}&53.74\textsubscript{$\pm$0.27}&70.71\textsubscript{$\pm$0.35}
  & 67.66\textsubscript{$\pm$0.63}&81.62\textsubscript{$\pm$0.34}&88.80\textsubscript{$\pm$1.43}&28.72\textsubscript{$\pm$1.70} &  64.56\textsubscript{$\pm$0.23}&58.38\textsubscript{$\pm$0.76}&66.97\textsubscript{$\pm$0.43}&74.23\textsubscript{$\pm$.001} \\

  Full FT  & 86.4M (100\%) & 70.57\textsubscript{$\pm$15.13}&76.21\textsubscript{$\pm$1.95}&57.14\textsubscript{$\pm$1.46}&66.90\textsubscript{$\pm$2.72} & 
  75.52\textsubscript{$\pm$2.89}&98.15\textsubscript{$\pm$0.56}&89.54\textsubscript{$\pm$1.88}&\textbf{59.63\textsubscript{$\pm$2.53}}
  &58.38\textsubscript{$\pm$0.64}&\textbf{80.67\textsubscript{$\pm$1.22}}&63.05\textsubscript{$\pm$0.85}&68.27\textsubscript{$\pm$0.43}  
  \\ \hline
 
\end{tabular}}
\label{tab:domainbed_original}
\end{table*}

\begin{table*}[t!]
\centering
\caption{TerraIncognita results (mean accuracy and standard deviation in $\%$). All methods were trained for 20,000 steps.}
\scalebox{1}{
\begin{tabular}{llccccc}
\hline
Method & \# Trainable Parameters  & \multicolumn{4}{c}{TerraIncognita} \\
 & (\% full ViT params)  & L100 & L38  & L43 & L46 \\
 &   &  (OOD) &  (ID) & (OOD) & (OOD) \\
\hline
 $\hat{B} A_{rand}$ $(r=8)$  & 0.16M-0.2M (0.18-0.29\%) & 16.59\textsubscript{$\pm$2.59}& 79.88\textsubscript{$\pm$0.45} &6.46\textsubscript{$\pm$1.25} &10.96\textsubscript{$\pm$0.52}  \\ 
 $\hat{B} A_{rand}$ $(r=16)$ & 0.3M-0.4M (0.36-0.46\%) & 14.14\textsubscript{$\pm$1.45}&80.48\textsubscript{$\pm$0.99}&7.74\textsubscript{$\pm$0.26}&11.09\textsubscript{$\pm$0.76}  \\
 $B_{rand}\hat{A}$ $(r=8)$ & 0.16M-0.2M (0.18-0.29\%) & 12.82\textsubscript{$\pm$0.84}&78.65\textsubscript{$\pm$0.57}&3.42\textsubscript{$\pm$0.81}&7.24\textsubscript{$\pm$1.36}  \\
$B_{rand}\hat{A}$ $(r=16)$  & 0.3M-0.4M (0.36-0.46\%) &  17.58\textsubscript{$\pm$1.01}&78.89\textsubscript{$\pm$0.55}&8.41\textsubscript{$\pm$1.88}&7.62\textsubscript{$\pm$0.56} \\ \hline
  LoRA $(r=8)$  & 0.3M-0.4M (0.35-0.46\%) & \textbf{41.36\textsubscript{$\pm$2.94}}&87.33\textsubscript{$\pm$.13}&13.48\textsubscript{$\pm$2.19}&7.76\textsubscript{$\pm$1.69}
  
  \\
  Linear Probing  & 0.004M (0.00\%) &  
13.82\textsubscript{$\pm$.20}&69.82\textsubscript{$\pm$0.36}&10.06\textsubscript{$\pm$.45}&13.90\textsubscript{$\pm$.49}  \\

  Full FT  & 86.4M (100\%) &
38.33\textsubscript{$\pm$6.50}&\textbf{95.05\textsubscript{$\pm$.31}}&\textbf{14.18\textsubscript{$\pm$2.33}}&\textbf{19.50\textsubscript{$\pm$1.53}}
  \\ \hline
 
\end{tabular}}
\label{tab:domainbed_extra}
\end{table*}

\begin{table*}[!t]
\centering
\caption{Generalization results (train set - test set accuracy in $\%$) for DomainBed.}
\scalebox{0.4}{
\begin{tabular}{lcccccccccccccccccc}
\hline
Method & \# Trainable Parameters & \multicolumn{4}{c}{VLCS}& \multicolumn{4}{c}{PACS}& \multicolumn{4}{c}{OfficeHome} & \multicolumn{4}{c}{TerraIncognita} \\
 & (\% full ViT params) & Caltech101 & LabelMe  & SUN09 & VOC2007 & Art & Cartoon  & Photo & Sketch  & Art & Clipart  & Product & Photo & L100 & L38 & L43 & L46  \\
 &   &  (OOD) &  (ID) & (OOD) & (OOD) &  (OOD) &  (ID) & (OOD) & (OOD)&  (OOD) &  (ID) & (OOD) & (OOD)&  (OOD) &  (ID) & (OOD) & (OOD) \\
\hline
 $\hat{B} A_{rand}$ $(r=8)$  & 0.2M-M (0.29-0.\%) & -1.72\textsubscript{$\pm$2.24} & 11.82\textsubscript{$\pm$1.21} & 28.09\textsubscript{$\pm$2.04} & 16.98\textsubscript{$\pm$0.74} & 15.82\textsubscript{$\pm$0.68} & 3.83\textsubscript{$\pm$0.70} & 0.83\textsubscript{$\pm$0.30} & 57.34\textsubscript{$\pm$0.89} & 15.94\textsubscript{$\pm$0.28} & 11.87\textsubscript{$\pm$1.14} & 11.51\textsubscript{$\pm$0.47} & 7.97\textsubscript{$\pm$0.56} & 64.20\textsubscript{$\pm$2.58} & 0.91\textsubscript{$\pm$0.43} & 74.33\textsubscript{$\pm$1.26} & 69.82\textsubscript{$\pm$0.53}   \\ 
 $\hat{B} A_{rand}$ $(r=16)$ & 0.3M-0.4M (0.36-0.46\%) & -2.48\textsubscript{$\pm$0.69} & 9.99\textsubscript{$\pm$1.44} & 28.11\textsubscript{$\pm$2.74} & 15.43\textsubscript{$\pm$0.70} & 12.92\textsubscript{$\pm$0.87} & 3.76\textsubscript{$\pm$0.40} & 0.22\textsubscript{$\pm$0.67} & 57.42\textsubscript{$\pm$0.62} & 16.22\textsubscript{$\pm$0.93} & 12.25\textsubscript{$\pm$1.23} & 11.81\textsubscript{$\pm$0.34} & 8.19\textsubscript{$\pm$0.87} & 66.62\textsubscript{$\pm$1.54} & 0.28\textsubscript{$\pm$1.18} & 73.02\textsubscript{$\pm$0.24} & 69.67\textsubscript{$\pm$0.56}  \\
 $B_{rand}\hat{A}$ $(r=8)$ & 0.2M-M (0.29-0.\%) & 0.19\textsubscript{$\pm$0.86} & 10.66\textsubscript{$\pm$0.86} & 27.48\textsubscript{$\pm$1.86} & 16.93\textsubscript{$\pm$0.19} & 19.79\textsubscript{$\pm$0.66} & 4.81\textsubscript{$\pm$0.99} & 4.78\textsubscript{$\pm$0.29} & 67.19\textsubscript{$\pm$1.34} & 17.73\textsubscript{$\pm$0.30} & 13.73\textsubscript{$\pm$0.86} & 12.08\textsubscript{$\pm$0.42} & 7.45\textsubscript{$\pm$0.65} & 65.86\textsubscript{$\pm$0.64} & 0.04\textsubscript{$\pm$0.60} & 75.27\textsubscript{$\pm$0.50} & 71.45\textsubscript{$\pm$1.17}  \\
$B_{rand}\hat{A}$ $(r=16)$  & 0.3M-0.4M (0.36-0.46\%) & -1.50\textsubscript{$\pm$2.88} & 9.75\textsubscript{$\pm$0.85} & 27.34\textsubscript{$\pm$2.07} & 16.97\textsubscript{$\pm$0.61} & 15.89\textsubscript{$\pm$0.96} & 3.44\textsubscript{$\pm$0.54} & 1.69\textsubscript{$\pm$0.30} & 62.30\textsubscript{$\pm$0.83} & 15.20\textsubscript{$\pm$0.53} & 13.07\textsubscript{$\pm$1.30} & 11.38\textsubscript{$\pm$0.38} & 6.53\textsubscript{$\pm$0.64} &  62.17\textsubscript{$\pm$1.41} & 0.86\textsubscript{$\pm$0.96} & 71.34\textsubscript{$\pm$1.91} & 72.13\textsubscript{$\pm$0.15} \\ \hline
  LoRA $(r=8)$  & 0.3M-0.4M (0.35-0.46\%) & 52.94\textsubscript{$\pm$1.48} & 24.03\textsubscript{$\pm$0.16} & 37.10\textsubscript{$\pm$3.25} & 33.28\textsubscript{$\pm$1.64}
    
  & 18.23\textsubscript{$\pm$0.74} & 4.70\textsubscript{$\pm$0.57} & 4.22\textsubscript{$\pm$0.43} & 49.74\textsubscript{$\pm$1.44}
  & 26.07\textsubscript{$\pm$0.39} & 17.97\textsubscript{$\pm$1.80} & 22.53\textsubscript{$\pm$0.63} & 17.57\textsubscript{$\pm$0.23}
  &47.53\textsubscript{$\pm$2.80} & 1.56\textsubscript{$\pm$0.24} & 75.41\textsubscript{$\pm$2.29} & 81.12\textsubscript{$\pm$1.73}
  
  \\
  Linear Probing  & 0.004M (0.00\%) &  
  -12.03\textsubscript{$\pm$2.11} & 3.04\textsubscript{$\pm$1.38} & 24.88\textsubscript{$\pm$0.47} & 7.91\textsubscript{$\pm$0.79}
  & 17.18\textsubscript{$\pm$0.13} & 3.22\textsubscript{$\pm$0.40} & -3.96\textsubscript{$\pm$1.90} & 56.13\textsubscript{$\pm$1.33} &  6.02\textsubscript{$\pm$0.21} & 12.20\textsubscript{$\pm$1.03} & 3.61\textsubscript{$\pm$0.51} & -3.65\textsubscript{$\pm$0.19} & 55.17\textsubscript{$\pm$0.28} & -0.82\textsubscript{$\pm$0.31} & 58.94\textsubscript{$\pm$0.52} & 55.10\textsubscript{$\pm$0.52}  \\

  Full FT  & 86.4M (100\%) & 29.03\textsubscript{$\pm$15.27} & 23.40\textsubscript{$\pm$2.05} & 42.47\textsubscript{$\pm$1.83} & 32.70\textsubscript{$\pm$2.27} & 
  24.41\textsubscript{$\pm$2.94} & 1.78\textsubscript{$\pm$0.54} & 10.38\textsubscript{$\pm$1.90} & 40.30\textsubscript{$\pm$2.49}
  &40.23\textsubscript{$\pm$0.48}& 17.94\textsubscript{$\pm$1.36}& 35.56\textsubscript{$\pm$1.02}& 30.35\textsubscript{$\pm$0.53}  &
  59.84\textsubscript{$\pm$6.53} & 3.12\textsubscript{$\pm$0.26} & 83.99\textsubscript{$\pm$2.31} & 78.67\textsubscript{$\pm$1.47}
  \\ \hline 
\end{tabular}}
\label{tab:domainbed_generalization}
\end{table*}

\end{document}